\documentclass[lettersize,journal]{IEEEtran}
\usepackage{amsmath,amsfonts}
\usepackage{algorithmic}
\usepackage{algorithm}
\usepackage{array}
\usepackage[caption=false,font=normalsize,labelfont=sf,textfont=sf]{subfig}
\usepackage{textcomp}
\usepackage{stfloats}
\usepackage{url}
\usepackage{verbatim}
\usepackage{graphicx}
\usepackage{cite}
\usepackage{amssymb}

\usepackage{booktabs}
\usepackage{multirow}
\usepackage{float}
\usepackage{enumitem}

\usepackage{soul}
\usepackage{color}
\usepackage{cite}

\usepackage{subfig}

\usepackage{tabularx}

\usepackage{arydshln}

\usepackage{orcidlink}
\usepackage{pifont}
\usepackage{stfloats}
\newcolumntype{C}{>{\centering\arraybackslash}X} % this creates a new type of column C which is centered and takes up the remaining space

\begin{document}

\title{Mixture of Enhanced-View Experts for Multi-Query Vehicle ReID and A Large-Scale Benchmark}

\author{Aihua~Zheng,
        Jie~Zhen,
        Chenglong~Li,
        Jiaxiang~Wang,
        Jin~Tang
% <-this % stops a space

\thanks{
% Manuscript received 12, June, 2024; revised 2 September 2024; accepted 20 October 2024. 
% Date of publication 2 September 2024; date of current version 7 December 2023.

% This research is partly supported by the National Natural Science Foundation of China (No. 62372003), the Natural Science Foundation of Anhui Province (No.2308085Y40 and No. 2208085J18) and the University Synergy Innovation Program of Anhui Province (No. GXXT-2022-036).

This research is supported in part by the National Natural Science Foundation of China under Grants 62372003 and 62376004, the University Synergy
Innovation Program of Anhui Province under Grant GXXT-2022-036, the Natural Science Foundation of Anhui Province under Grants 2308085Y40
and 2208085J18, and the Open Research Project of the Anhui Provincial Key Laboratory of Security Artificial Intelligence under Grant SAI202401. (* The
corresponding author is Chenglong Li.)
}

\thanks{Aihua Zheng and Chenglong Li are with Information Materials and Intelligent Sensing Laboratory of Anhui Province, Anhui Provincial Key Laboratory of Multimodal Cognitive Computation, School of Artificial Intelligence, Anhui University. Hefei, 230601, China (e-mail: ahzheng214@foxmail.com; lcl1314@foxmail.com.}
\thanks{Jie Zhen is with Anhui Provincial Key Laboratory of Multimodal Cognitive Computation, School of Artificial Intelligence, Anhui University. Hefei, 230601, China (e-mail: wa23201033@stu.ahu.edu.cn).}
\thanks{Jiaxiang Wang is with the School of Artificial Intelligence, Anhui University of Science and Technology, Hefei 232001, China (e-mail: Netizenwjx@foxmail.com).}

\thanks{Jin Tang is with Anhui Provincial Key Laboratory of Multimodal Cognitive Computation, School of Computer Science and Technology, Anhui University. Hefei, 230601, China (e-mail: tangjin@ahu.edu.cn).}
}

\maketitle

\begin{abstract}
%Single-query vehicle re-identification (ReID) faces great challenges in maintaining robustness within complex visual scenarios. In contrast, multi-query vehicle ReID utilizes complementary information from diverse views for robust feature learning. 
Multi-query vehicle ReID aims to leverage complementary information from diverse views for robust feature learning. However, current methods suffer from simplistic feature fusion and thus easily ignores some important view information and cross-view relationships. 
To handle these problems, this work presents a novel approach called Mixture of Enhanced-View Experts (EV-MoE), which enhances the feature representation of each view and efficiently integrate the view-specific enhanced features by MoE, for robust multi-query ReID.
In particular, we design a mixture of enhanced-view experts module, which consists of two parts including view-specific feature enhancement sub-Module (VFEM) and dynamic multi-view fusion sub-Module (DMFM). Moreover, we further introduce Multi-view Alignment Loss (MAL), which aligns features through bidirectional cross-view contrastive learning and reconstruction constraints, addressing the challenges of consistency between multi-query features and single-image features. In addition, to evaluate multi-query ReID in real-world environments, we collect LCRI-1K, a large-scale vehicle ReID dataset with 1,090 identities, 107,805 images, across 23,637 cameras, where each vehicle appears in an average of 67.5 cameras, providing a comprehensive benchmark to test the robustness in complex environments. Extensive experiments demonstrate the robustness of CAFNet in addressing the multi-query vehicle ReID problem. The code is available at \url{https://github.com/xiaozhen28/CAFNet.}
\end{abstract}


\begin{figure}[t]
    \centering
    \includegraphics[width=1\columnwidth]{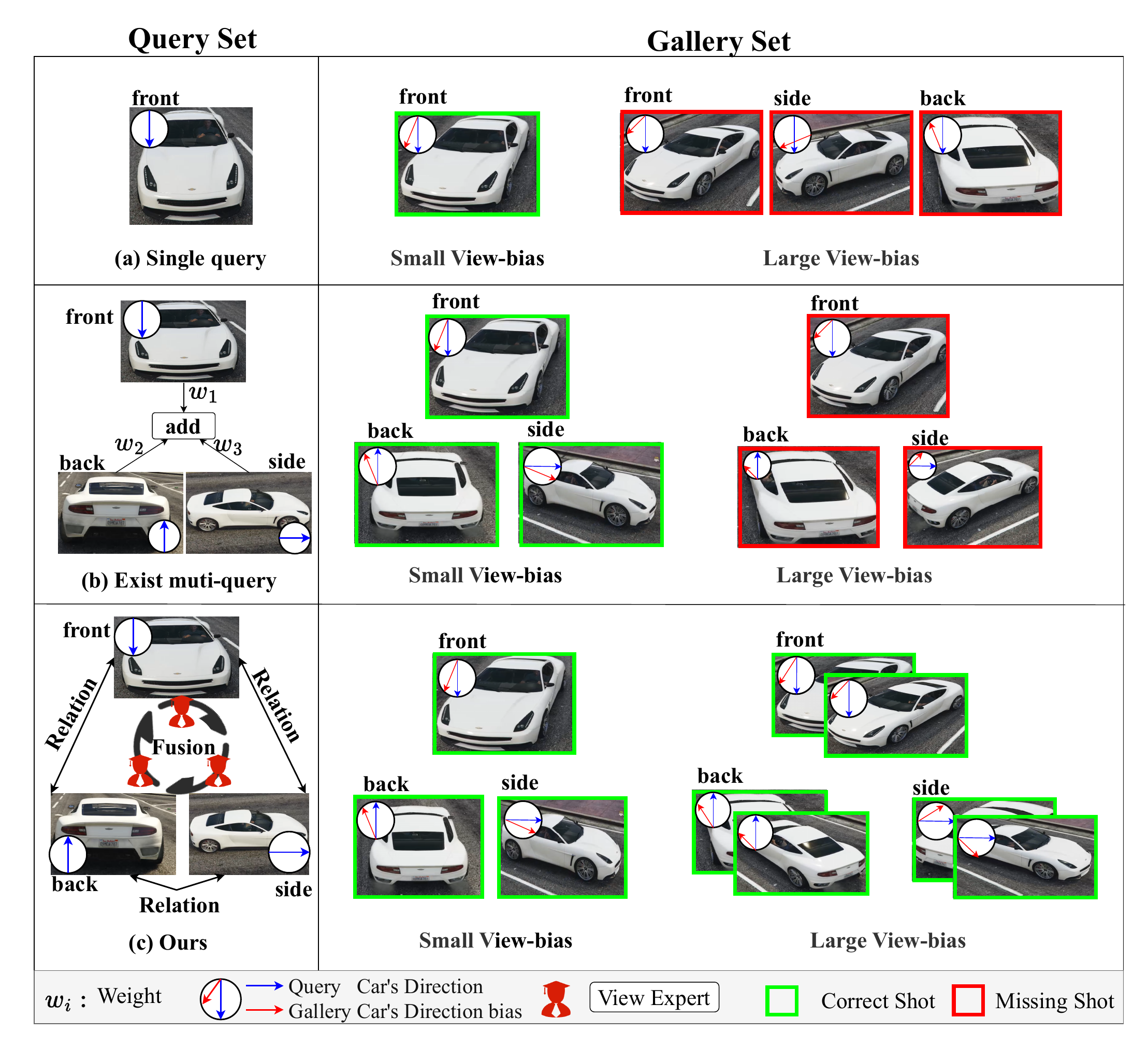} 
    \vspace{-4mm}
    \caption{Diagrams of motivation. This diagram compares three approaches: (1) Single-query methods show strong view dependency; (2) Current multi-query methods based on weighted feature fusion exhibit significant performance limitations under large view-bias conditions; (3) Our dynamic fusion intelligently combines complementary features by different view experts, achieving both view consistency and high discriminability.
    Green boxes indicate target IDs, while red boxes denote non-target IDs.
    }
    \label{fig:dongji}
\end{figure}

\section{Introduction}

\IEEEPARstart{V}{ehicle} re-identification (ReID) is vital for intelligent transportation systems, aiming to retrieve vehicles across non-overlapping cameras. While widely applied in urban surveillance, most image-based methods rely on single-query inputs, struggling with cross-viewpoint appearance variations (Fig.~\ref{fig:dongji}(a)).  
To address viewpoint diversity, prior work uses local feature alignment via vehicle keypoints~\cite{wang2017orientation,moskvyak2021keypoint,liu2018ram,liurTipRegin} or meta-information like attributes~\cite{li2021attributes} and spatio-temporal cues~\cite{liu2017provid}. However, single-query methods are limited: each image represents a specific viewpoint, failing to robustly match gallery images from diverse angles.  
Multi-view input strategies~\cite{jin2020uncertainty,ZhouMvinf} attempt to capture cross-view information but rely on single-query inference, restricting multi-query data utilization. This highlights the need for adaptive multi-view fusion to handle dynamic viewpoint changes.

% 多查询符合真实交通场景
% 为了单查询网络中视角差异问题，zheng等人首次提出来车辆多查询重识别方法
To address the challenge of viewpoint variations in single-query networks, Zheng~\emph{et al.}~\cite{zheng2023multi} pioneer the innovative multi-query vehicle ReID approach. Multi-query vehicle ReID means using multiple images of a car (from different viewpoint or cameras) together as the query, while the gallery set still stores single image for matching. By integrating complementary information from multi-query image scenarios, we can learn more comprehensive appearance feature representations of a target vehicle, which is expected to significantly mitigate the dramatic appearance variations caused by viewpoint differences. 
%% 条查询提出
% 多查询车辆重新识别就是通过使用来自不同场景/视角的一辆车的多张照片作为查询，同时gallery set 依然为单张图片。
% 传统多查询方法直接通多对多个单查询特征(或者分数)求均值，现有针对多查询提出的方法通过结合视角信息训练网络得到权重加权求和得到最终匹配结果，相对于单查询能够得到包含多个视角信息，但是在简单均值求和过程中存在细节丢失和视角偏置的问题，也没有充分探索视角间丰富关联特征,对新颖视角的匹配常常失败。
However, as shown in Fig.~\ref{fig:dongji}(b), existing multi-query vehicle ReID approaches typically combine multi-view information through either \textbf{feature/score averaging} or \textbf{viewpoint-weighted} strategies.
Averaging-based methods, which merely compute the mean of single-query features during testing, tend to lose fine-grained details and introduce view-bias due to their oversimplified aggregation.
Viewpoint-weighted approach (i.e., VCNet~\cite{zheng2023multi}) leverages viewpoint annotations to construct perspective-aware branch. While this mitigates small view-bias bias, its real-world applicability is limited due to the scarcity of viewpoint labels in large-scale scenarios. More critically, such frameworks inadequately model inter-view correlations, leading to poor generalization to large view-bias.

% 综上所述，如何充分发掘多个查询图片的之间的相关关系并且利用好丰富的细节信息得到更全面完整的融合特征成为多查询任务的关键。
% 同时，根据我们的调查研究，多查询reid面临新的挑战，可以为如下表述。

% \textbf{The core challenge in multi-query ReID tasks lies in comprehensively exploring inter-image correlations and utilizing rich detailed information to generate more robust fused features.} Based on our investigation, multi-query vehicle ReID faces several emerging challenges that can be summarized as follows: \textbf{(\romannumeral1) Viewpoint-invariant feature learning}: how to learn robust representations that maintain consistency across significantly different viewing angles, addressing fundamental appearance variations. \textbf{(\romannumeral2) Multi-view information fusion}: how to efficiently fuse multi-view information while avoiding redundancy and noise interference is another major challenge for multi-query ReID. \textbf{(\romannumeral3) Loss function design}: existing loss functions (i.e., ID loss, Triplet loss) are primarily designed for single-query scenarios and lack explicit constraints for multi-view features. \textbf{(\romannumeral4) Dataset limitations}: existing vehicle ReID datasets exhibit significant limitations in both scale and diversity. These constraints critically impair the validation of multi-query vehicle ReID methods. The field therefore urgently requires establishing larger-scale benchmark datasets with richer scenario variations and more comprehensive spatiotemporal coverage to facilitate practical deployment of vehicle ReID technologies.

% 数据集 样本展示
\begin{figure}[]
  \centering
  \includegraphics[width=1.0\linewidth]{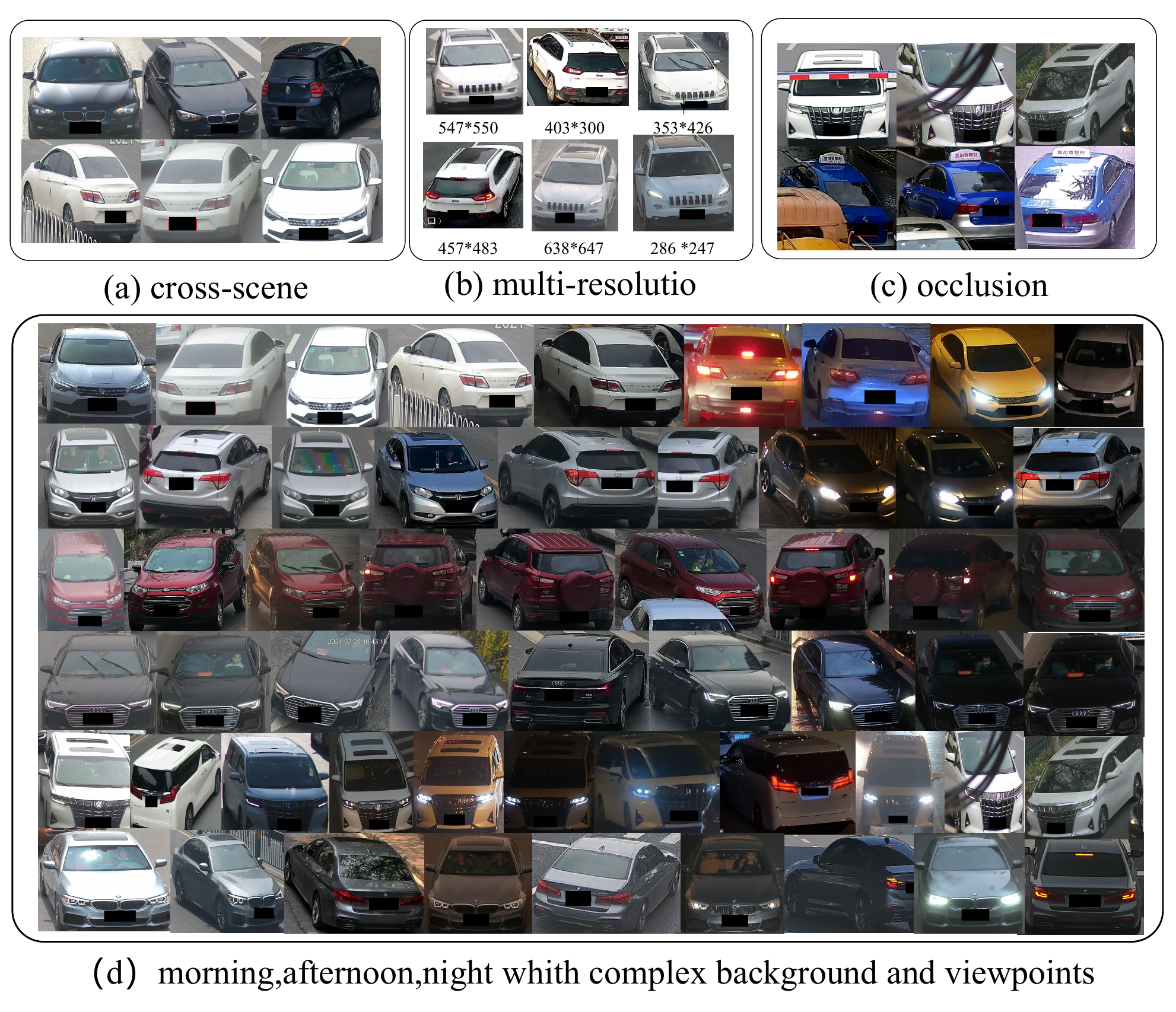}
  \vspace{-4mm}
  \caption{Challenges of LCRI-1K dataset. Each row represents a image of the same car under a different camera.
   (a) cross-scene;
   (b) multi-resolution;
   (c) occlusion;
   (d) morning,afternoon,night;
   The images in the same row indicate the same vehicle ID.}
   \label{fig:challeng}
\end{figure}

The core challenge in multi-query ReID tasks lies in comprehensively exploring inter-image correlations and utilizing rich detailed information to learn more robust fused features. Multi-query vehicle ReID faces challenges in viewpoint feature learning, cross-view fusion, loss function adaptation. In addition, existing vehicle ReID datasets exhibit significant limitations in both scale and diversity. These constraints critically impair the validation of multi-query vehicle ReID methods.
This paper proposes a Cross-view Adaptive Fusion Network (CAFNet) to address viewpoint differences and complementary information fusion challenges in multi-query ReID. CAFNet integrates multi-query features through Mixture of Enhanced-View Experts (EV-MoE), providing a robust baseline for multi-query vehicle ReID, as illustrated in Fig.~\ref{fig:dongji}(c).

%TOPREID 基于交叉注意力机制，对全局语义信息轮转方式注意到不同模态信息实现多模态信息。
% Wang~\emph{et al.}~~\cite{wang2024top} propose the Token Permutation Module (TPM) with a cyclic token permutation mechanism to achieve the spatial feature alignment among different image spectra and the effective aggregation of heterogeneous features. Existing token permutation module (TPM)~\cite{wang2024top} with cyclic token alignment lack explicit cross-view relation learning, limiting their effectiveness in multi-query scenarios. 
% VFEM子模块

Current view feature extraction methods (e.g., cross-attention~\cite{chen2021crossvit}, TPM~\cite{wang2024top}) significantly compromise feature discriminability in multi-query learning scenarios due to the lack of explicit cross-view relation. To explore inter-image correlations and address the uncontrolled vehicle viewpoints, we introduce a View-specific Feature Enhancement sub-Module (VFEM) as a critical component of our Mixture of Enhanced-View Experts (EV-MoE). VFEM is designed to enhance the feature representation of each view by leveraging complementary information from other viewpoints, enabling the network to adaptively learn discriminative features even when vehicle orientations are uncontrollable. By modeling inter-image correlations across different perspectives, VFEM strengthens the feature extraction capability of each expert, ensuring robust performance in multi-view vehicle ReID scenarios.
% View-specific Feature Enhancement sub-Module (VFEM) leverages cross-attention guided by the global information of the current viewpoint to extract and inject complementary local features from other viewpoints, enhancing the current viewpoint's specific representation while preserving its original local features. This approach mitigates the impact of image quality variations and occlusions, generating robust features with both viewpoint invariance and discriminative detail.

% 融合策略
To leverage rich multi-query information and learn robust fused representations, we propose a Dynamic Multi-view Fusion sub-Module (DMFM) as a critical component of our Mixture of Enhanced-View Experts (EV-MoE),  which adaptively fuses multi-view features through an expert network and an attention-driven gating mechanism. Unlike existing method (i.e., VAF~\cite{zheng2023multi}) that struggles to adapt to dynamic viewpoint changes in complex scenes due to reliance on static weight assignment or feature splicing, DMFM employs an expert network combined with an attention-driven gating mechanism to adaptively fuse multi-view features. This design enables the module to dynamically weight features from diverse viewpoints, prioritizing discriminative regions while suppressing irrelevant information. By leveraging enhanced view-specific features from VFEM, DMFM generates compact and invariant representations that effectively bridge cross-view discrepancies, enhancing the network's capability to integrate diverse viewpoints in multi-query ReID scenarios.
% This design enables the module to dynamically weight features from diverse viewpoints, prioritizing discriminative regions while suppressing irrelevant information. By leveraging enhancement view-specific feature from VFEM, DMFM generates compact and invariant representations that effectively bridge cross-view discrepancies, enhancing the network's capability to combine diverse viewpoints in multi-query ReID scenarios.
% DMFM includes three key innovations: a set of independent expert networks, enhanced by residual connections to ensure stable feature representation; a dynamic routing network that assigns expert weights based on feature similarity, using Softmax normalization and a learnable residual scaling factor for robust fusion; a computational network that explicitly models semantic associations between viewpoints, dynamically adjusting their contributions through an attention mechanism. This approach effectively resolves inconsistent feature distribution across viewpoints and large view-bias.

% 损失函数
To address inconsistent feature distribution in multi-query vehicle ReID, this paper proposes a Multi-view Alignment Loss (MAL). While existing loss functions (e.g., Cross-Entropy~\cite{mao2023cross}, Triplet~\cite{schroff2015facenet}, Center~\cite{wen2016discriminative} losses) effectively optimize intra-class and inter-class relationships for single images, they fail to maintain feature consistency in multi-query scenarios, significantly degrading cross-view retrieval performance.
The Multi-view Alignment Loss (MAL) unifies bi-directional contrastive learning and feature reconstruction. This framework simultaneously enforces uniform latent space distribution between multi-query features and single-query features while preserving view-specific details via MSE constraints, effectively mitigating viewpoint-induced retrieval errors through enhanced cross-view consistency.
% MAL combines contrastive learning with feature reconstruction constraints to align and preserve multi-query feature consistency. It employs a bi-directional contrastive learning framework to uniformly distribute fused and single-query features in the latent space, while using Mean Square Error (MSE) loss to ensure fused features reconstruct single-view features and retain key details. By balancing contrastive learning and reconstruction with adjustable weights, MAL enhances cross-view feature similarity and reduces retrieval errors caused by viewpoint differences.

At last, to address the lack of large-scale cross-scenario benchmarks, we propose LCRI-1K  the first city-level multi-query vehicle ReID dataset. Although existing vehicle ReID datasets provide important benchmarks to evaluate the state-of-the-art methods, their crucial issues lie in the restricted number of cameras, insufficient viewpoint diversity, and narrow spatio-temporal coverage.
% the crucial issue is the number of cameras is limited 
%(e.g., 12 cameras in VechileID~\cite{liu2016deep_vehiclesID}, 20 cameras in VeRI-776~\cite{liu2016large}, 174 cameras in VERI-Wild~\cite{lou2019veri}, 6142 cameras in MURI~\cite{zheng2023multi}). 
LCRI-1K dataset is built on the basis of urban intelligent traffic monitoring system, realizing full spatial coverage at the city level. 
It provides complete viewpoints (front/side-front, side, rear/side-rear) for vehicles. Additionally, it contains 107,805 images covering cross-scene recordings of 1,090 vehicle under 23,637 cameras, with an average of 67.5 heterogeneous scenes spanning each vehicle, and the time span covering continuous recordings throughout the year at all times of the day and night to completely capture seasonal changes, as shown in Fig.~\ref{fig:challeng}.

% 总结
Overall, the contributions of this paper are systematically summarized as follows:
\begin{itemize}

% \item We propose a Cross-view Adaptive Fusion Network (CAFNet) that integrates the Mixture of Enhanced-View Experts (EV-MoE) and Multi-view Alignment Loss (MAL). The EV-MoE enhances view-specific features and fuses them dynamically. MAL reduces cross-view differences. This enables CAFNet to aggregate multi-view features, learn unified representations, and boost multi-query vehicle ReID performance.

%%%MoE模块

% \item A novel Cross-view Adaptive Fusion Network (CAFNet) is proposed, integrating the  Mixture of Enhanced-View Experts (EV-MoE) (include View-specific Features Enhancement sub-Module (VFEM) and Dynamic Multi-view Fusion sub-Module (DMFM)), and Multi-view Alignment Loss (MAL). CAFNet dynamically aggregates multi-view features, prioritizes discriminative regions, and learns a unified global representation, significantly improving multi-query vehicle ReID performance.

% \item To address the inefficiency of simplistic feature fusion in multi-query ReID, we propose the EV-MoE framework to achieve adaptive enhancement and fusion of view-specific features via attention mechanisms and Mixture of Experts.

% \item To address the inefficiency of simplistic feature fusion in multi-query ReID, we propose the EV-MoE framework, which adaptively enhances and dynamically fuses via attention mechanisms and Mixture of Experts.

% \item To fully exploit complementary information from diverse views and leverage the relationships between them, we propose EV-MoE, which uses attention mechanisms and Mixture of Experts for view-specific feature enhancement and dynamic fusion.

\item To leverage the inter-view correlations for view-specific feature enhancement and exploit complementary information from diverse views for dynamic fusion, we propose EV-MoE, which employs attention mechanisms and Mixture of Experts for view-specific feature enhancement and dynamic fusion.

%%%损失函数
% \item To resolve semantic inconsistency in multi-query features, we introduce the MAL, which enforces feature alignment via cross-view contrastive learning and reconstruction constraints to bridge discrepancies between multi-query and single-image representations.
% \item We propose MAL, which employs bidirectional cross-view contrastive learning and reconstruction constraints to enforce semantic consistency between multi-query features and single-image features. This mechanism resolves the challenge of multi-view feature alignment, ensuring that fused features retain both cross-view complementarity and single-view discriminability.

% \item To resolve semantic inconsistency in multi-query features, we introduce the MAL loss, enforcing feature alignment via contrastive learning and reconstruction constraints.

% \item To resolve semantic inconsistency in multi-query features, we introduce the MAL, which enforces feature alignment via cross-view contrastive learning and reconstruction constraints to bridge discrepancies between multi-query and single-image representations.

% \item In the inference stage of multi-query ReID, semantic misalignment between multi-query and single-view features severely hinders matching accuracy. We introduce the MAL, which enforces feature alignment via cross-view contrastive learning and reconstruction constraints to bridge discrepancies between multi-query and single-image representations.

\item To resolve semantic inconsistency in multi-query features, we introduce the MAL, which enforces feature alignment via cross-view contrastive learning and reconstruction constraints to bridge discrepancies between multi-query and single-image representations.

%%% 数据集
% \item We introduce a large-scale vehicle ReID dataset with multi-view urban images collected over a year. It addresses the lack of high-quality datasets for cross large number of cameras multi-query scenarios and will be publicly released.
% \item  We introduce LCRI-1K, a large-scale cross-scene vehicle ReID dataset with multi-view images collected over one year in urban environments. It fills the gap for high-quality datasets in multi-camera multi-query scenarios and will be publicly released to promote research.

\item We introduce a large-scale vehicle ReID dataset with multi-view urban images collected over a year. It addresses the lack of high-quality datasets for cross large number of cameras multi-query scenarios and will be publicly released.

%%% 实验结果
% \item CAFNet demonstrates remarkable performance on both the MURI~\cite{zheng2023multi} and our LCRI-1K benchmark, outperforming outperforming existing methods in key metrics (mAP, mCSP, mINP, Rank1). Its robust performance in complex scenarios validates its effectiveness in handling multi-query differences in real environments.

\item CAFNet shows remarkable performance on MURI~\cite{zheng2023multi} and LCRI-1K, outperforming existing methods in mAP ($\uparrow$3.0\%), mCSP ($\uparrow$1.1\%), mINP ($\uparrow$3.6\%) and Rank1 ($\uparrow$0.7\%) on LCRI-1K. Its robust performance in complex scenarios proves its effectiveness in handling multi-query differences in real environments.

\end{itemize}

\section{Related Work}
\label{sec:related}
We briefly reviews the related work in the following two fields Vehicle ReID methods and Vehicle ReID Datasets.

\subsection{Vehicle ReID Methods}
Most vehicle ReID methods depend on a single-query image \cite{liu2016deep_vehiclesID,chen2020orientation,li2020vehicle,lou2019embedding}, which limits their ability to differentiate similar vehicles. To improve feature learning, several approaches have been introduced.
% Region of Interest and Attention Mechanisms
Some studies\cite{zhou2018aware} have utilized region-of-interest and attention mechanisms to enhance the learning of detailed features required to distinguish vehicles of similar appearance.
He~\emph{et al.}~\cite{he2019part} propose a partial regularization method that identifies regions of interest using pre-trained detectors, enhancing the model's ability to perceive subtle differences. 
Zhang~\emph{et al.}~\cite{an2019part} develop an attention network that emphasizes important local regions, while Khorramshshi~\emph{et al.}~\cite{khorramshahi2019dual} focus on key-point features based on orientations.

% Leveraging Additional Annotation 
Some studies propose to identify subtle differences between different vehicles by leveraging additional annotation information in the dataset.
%zheng等人通过Leveraging Additional Annotation实现更有效的车辆重识别
Zheng~\emph{et al.}~\cite{li2022attribute} achieve more effective vehicle ReID through attribute enhancement methods. Liu~\emph{et al.}~\cite{liu2017provid} use multi-modal data from surveillance to refine searches, and Wang~\emph{et al.}~\cite{wang2017orientation} extract local features using keypoint locations.
Zheng~\emph{et al.}\cite{li2022attribute} fuse camera views and vehicle characteristics to enrich features.
% 一些研究通过在查询中挖掘更多信息来提升车辆重识别的性能。 通过车辆解析学习判别性部分级特征，并建模部件之间的相关性，实现精确的部分对齐。
Some studies have improved the performance of vehicle ReID by mining more information in the query.
Jin~\emph{et al.}~\cite{jin2020uncertainty} explore multi-shot images in a teacher-student framework but still rely on single-image information during inference. Liu~\emph{et al.}~\cite{liu2020beyond} learn discrimination between different IDs.
Chu~\emph{et al.}~\cite{chu2019vehicle} learn two metrics for similar viewpoints and different viewpoints in two feature spaces.
% 通过视角解析和共同可见视角注意力机制，实现了细粒度的特征对齐与增强
%Zheng等人提出了一种通过预训练的视角判定网络，自适应地结合来自不同车辆视角的互补信息，从而实现多查询车辆重识别的方法。

Zheng~\emph{et al.}~\cite{zheng2023multi} propose a method to achieve multi-query vehicle ReID by adaptively combining complementary information from different vehicle perspectives through a pre-trained view determination network. Meng~\emph{et al.}~\cite{meng2020parsing} propose perspective parsing and co-visible perspective attention mechanisms to achieve fine-grained feature alignment and enhancement.
Through combining region-specific and cross-level feature and graph structure relationship~\cite{zhao2021heterogeneous} mitigates perspective invariance.

\subsection{Vehicle ReID Datasets}
% 现有的车辆重识别（Re-ID）方法主要基于个公共数据集进行评估：VeRI-776、VehicleID 、VERI-Wild、MURI。VehicleID 数据集则包含26,267辆汽车的221,763张图片，主要包括正面和背面视角。
Existing vehicle ReID methods are mainly evaluated based on four public datasets: VeRI-776~\cite{liu2016large}, VehicleID~\cite{liu2016deep_vehiclesID}, VERI-Wild~\cite{lou2019veri} and MURI~\cite{zheng2023multi}. 
% VeRI-776 数据集包含776辆汽车的49,360张图片，这些图片由20个摄像头在1.0平方公里的区域内短时间拍摄，每辆车至少出现在2个、最多18个不同的摄像头视角中。
The VeRI-776~\cite{liu2016large} dataset contains 49,360 images of 776 cars, which are taken by 20 cameras in an area of 1.0 square kilometers for a short period of time (4:00 PM to 5:00 PM during the
day), with each vehicle being captured by at least 2 and at most 18 cameras. Each car appears in at least 2 and at most 18 different camera views. 
The VehicleID~\cite{liu2016deep_vehiclesID} dataset contains 221,763 images of 26,267 cars, mainly from front and back views. VehicleID~\cite{liu2016deep_vehiclesID} divides the test set into 3 subsets, large, medium and small, according to the size of the vehicle images. VehicleID~\cite{liu2016deep_vehiclesID} contains limited views (only two views (i.e.,front view and rear view). In addition, the images in this
dataset mainly contain less complex backgrounds, occlusions and illumination changes.
% VERI-Wild 数据集收集于200平方公里的郊区，包含416,314张图片，涉及40,671辆汽车，由174个交通摄像头拍摄。
VERI-Wild~\cite{lou2019veri} is collected in a 200 Square Kilometers suburban areas and contains 416,314 images of 40,671 vehicles taken by 174 traffic cameras. The training set consists of 277,797 images (30,671 IDs) and the testing set consists of 138,517 images (10,000 IDs). Similarly, the testing set of VERI-Wild~\cite{lou2019veri} is divided into three subsets according to image size: large, medium, and small. The vehicle images in VERI-Wild~\cite{lou2019veri} mainly have little variability in views, mostly in front and rear views.
MuRI~\cite{zheng2023multi} contains 200 vehicle IDs captured by 6142 cameras from a real-life transportation surveillance system covering over 1000 $km^2$ urban area. Each vehicle in MuRI~\cite{zheng2023multi} crosses 34.6 cameras in average, varying from 30 to 50 cameras.

Despite the results achieved on these datasets, vehicle ReID still has problems in real-world scenarios. These datasets contain only a small number of cameras and scenes. VeRI-776~\cite{liu2016large}, VehicleID~\cite{liu2016deep_vehiclesID}, VERI-Wild~\cite{lou2019veri} and MuRI~\cite{zheng2023multi} are captured by 20, 12, 174 and 6142 cameras, respectively, which is far less than the tens of thousands of cameras in smart cities. At the same time, the vehicle perspective is mainly concentrated on the front and back, lacking side images, and the number of cameras that each vehicle passes through is limited, which affects the evaluation of cross-scene retrieval capabilities.

% 网络主图
\begin{figure*}[t]
  \centering
    \resizebox{1\textwidth}{!}{
    \includegraphics[width=1\linewidth]{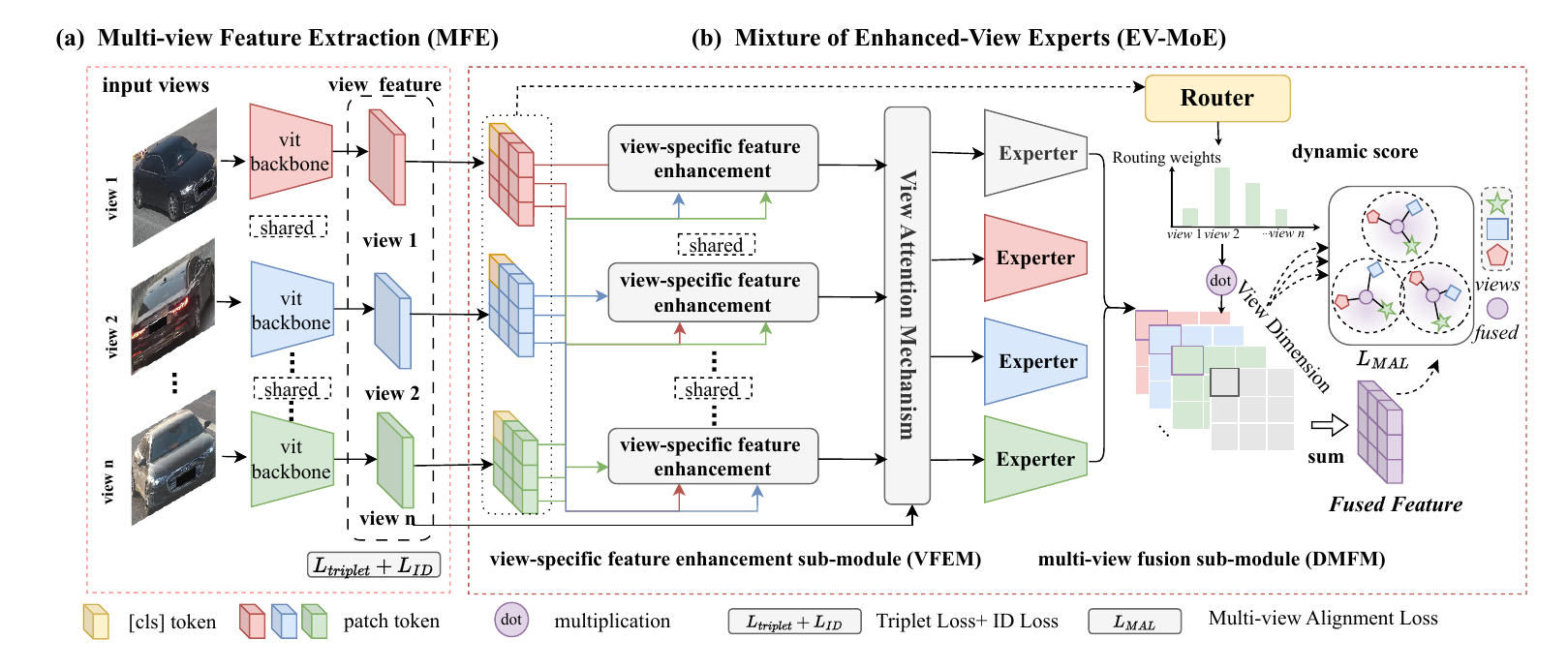} 
    }
       \caption{An illustration of our propose Cross-view Adaptive Fusion Network (CAFNet).
       CAFNet uses the Multi-view Feature Extraction (MFE) module to extract features from various input views, while applying id and triplet losses for constraint. Then, the Mixture of Enhanced-View Experts (EV-MoE) module, with its View-specific Feature Enhancement sub-Module (VFEM) and Multi-view Fusion sub-Module (DMFM), enhances and fuses these features. Finally, the Multi-view Alignment Loss (MAL) regularizes the fused features, allowing CAFNet to produce a unified representation and boost multi-query ReID performance. 
       % First, features from different input views are extracted by Multi-view Feature Extraction (MFE). Then, the View-specific Feature Enhancement sub-Module (VFEM) is utilized to obtain view-enhancement features.
       % Meanwhile, features from different input views are constrained by id loss and triplet loss.
       % Finally, dynamic multi-view fusion sub-Module (DMFM) dynamically fuses the view-enhancement features to obtain a unified feature representation. The whole framework can obtain more discriminative features for multi-query ReID.
       }
  \label{fig:Overall}

\end{figure*}

\section{Method} 
\label{sec:methods}
% 在这项工作中，为了有效结合来自不同车辆视角的互补信息，我们提出了一种基于ViT的多查询车辆再识别的网络。Vehicle Multi-query Transformer for ReIDNetwork
%如图 2 所示，我们提出的 CAFNet 由四个关键部分组成： 视角特定特征提取（Shared Feature Extraction）、跨视角感知（Cross-View Perception Module ,VFEM）和多视角特征融合（muti-view Dynamic Fusion Module，mdf）。此外，我们还加入跨视角约束，以进一步增强模型对多视角感知能力。我们将在以下几个小节中介绍这些关键模块。
%如图所示，我们提出的 CAFNet 由三个关键部分组成：共享特征提取、跨视角感知（VFEM）模块和多视角动态融合（mdf）模块。此外，我们还引入了跨视角约束，以进一步增强多视角感知模型。接下来，我们将在以下小节中详细介绍这些关键模块。
% In this work, we propose a ViT-based~\cite{alexey2020image} multi-query vehicle ReID network (CAFNet) to effectively combine complementary information from different vehicle viewpoints. This network 
% presents a novel approach called Mixture of Enhanced-View Experts (EV-MoE), which enhances the feature representation of each view and efficiently integrate the view-specific enhanced features by MoE for robust multi-query ReID. In particular, we design a mixture of enhanced-view experts module, which consists of two parts including view-specific feature enhancement sub-Module (VFEM) and dynamic multi-view fusion sub-Module (DMFM). leverages view-specific feature enhancement sub-Module (VFEM) and dynamic multi-view fusion sub-Module (DMFM) sub-Module to extract and integrate features from various perspectives, achieving a more comprehensive and robust feature representation that significantly enhances ReID performance.
In this work, we propose a Vision Transformer (ViT) based~\cite{alexey2020image} multi-query vehicle re-identification (ReID) network named CAFNet, which is designed to effectively combine complementary information from different vehicle viewpoints. At the core of CAFNet lies a novel approach, the Mixture of Enhanced-View Experts (EV-MoE), which enhances the feature representation of each view and efficiently integrates the view-specific enhanced features via the MoE mechanism to enable robust multi-query ReID. The EV-MoE comprises two essential sub-Modules: the View-specific Feature Enhancement sub-Module (VFEM) and the Dynamic Multi-view Fusion sub-Module (DMFM). VFEM adaptively boosts the discriminative features of each vehicle view while preserving identity-related characteristics, while DMFM dynamically weights and merges the enhanced features output by VFEM from multiple views. By jointly extracting and fusing features across various perspectives, these two sub-Modules collaborate to form a more comprehensive and robust feature representation, thereby significantly improving the ReID performance.
As shown in the Fig.~\ref{fig:Overall}, our proposed CAFNet consists of four key components: Multi-view Feature Extraction (MFE), View-specific Features Enhancement sub-Module (VFEM), Dynamic Multi-view Fusion sub-Module, and Multi-view Alignment Loss (MAL). Next, we describe these key modules in detail in the following subsections.

\subsection{Multi-view Feature Extraction (MFE)}
% 为了在提取不同了视角特定的特征的同时缓解视角差异的影响，我们为多视角态输入部署了共享视觉转换器（ViT）。
As show in Fig.~\ref{fig:Overall}(a), in order to mitigate the effects of viewpoint differences while extracting different viewpoint-specific features, we deployed a shared Visual Transformer (ViT)~\cite{alexey2020image} for multi-view state inputs, can be formulated as:
\begin{equation} 
  F_{1} = \mathrm{ViT}(I_{1}),
  F_{2} = \mathrm{ViT}(I_{2}),
  ...,
  F_{n} = \mathrm{ViT}(I_{n}),
\end{equation}
\noindent where $I_1$, $I_2$ and $I_n$ represent the different input images of a same vehicle , respectively. The tokenized features $F_1$, $F_2$ and $F_n$, each of which has a shape of $\mathbb{R}^{D \times (N_{p}+1)}$, are extracted before the last layer of $\mathrm{ViT}$.

In contrast to the conventional single-query vehicle ReID, multi-query vehicle ReID exhibits significant paradigm differences during the testing phase: the query set requires the aggregation of features from multiple images, whereas the gallery set remains characterized by single image feature extraction. This inconsistency may lead to a degradation in the model's generalization capability. To address this issue, this paper proposes a progressive training strategy: initially, the model is trained with the Eq.~\eqref{eq:loss_reid} $L_{reid}$ to enhance single image feature learning, followed by the introduction of dynamic weighted fusion to simulate multi-query feature generation. This approach ensures consistency between the training and testing phases, thereby improving the model's robustness. The $L_{reid}$ can be formulated as:
% id loss
\begin{equation}
    \mathcal{L}_{id}=-\frac{1}{N}\sum_{i=1}^{N}log(p(y_i|x_i)),
\end{equation}
where \( N \) is the number of samples, and \( p(y_i | x_i) \) denotes the probability that the feature \( x_i \) belongs to the correct identity \( y_i \).
% triplet loss
\begin{equation}
    \mathcal{L}_{triplet} = \max(0, d(a, p) - d(a, n) + \alpha),
\end{equation}
where \( a \), \( p \), and \( n \) represent the anchor, positive, and negative samples. Respectively, \( d(\cdot, \cdot) \) is the distance metric and \( \alpha \) is a margin parameter.
% reid loss
\begin{equation}
\label{eq:loss_reid}
    \mathcal{L}_{reid} = \lambda_{id} \mathcal{L}_{id} + \lambda_{triplet} \mathcal{L}_{triplet},
\end{equation}
where \(\lambda_{\text{id}}\) and \( \lambda_{\text{triplet}} \) are weighting coefficients that control the influence of each component.

\subsection{View-specific Feature Enhancement sub-Module (VFEM)}
\label{method:VFEM}
%车辆再识别面临着不同摄像头视角下的巨大类内差异，而传统的融合方法往往会产生次优的特征表示。一个用于提取视点特定的专门模块对于弥合跨视角差异，同时保留识别特征至关重要。
Vehicle ReID faces substantial intra-class variations across camera views, where conventional fusion methods often yield suboptimal feature representations. A specialized module for extracting enhanced view-specific features is crucial to bridge cross-view discrepancies while preserving discriminative identity characteristics. To effectively exploit View-specific information across multi-view vehicle images, we propose the View-specific Feature Enhancement sub-Module (VFEM). This module takes the output features from the vision transformer as input and achieves cross-view feature enhancement through a hierarchical attention mechanism. The core processing pipeline Fig.~\ref{fig:Overall} incorporates the VFEM  sub-Module to extract enhanced view-specific features representations through cross-view feature correlation and transformation.

%cvp module display
\begin{figure}[]
    \includegraphics[width=1\columnwidth]{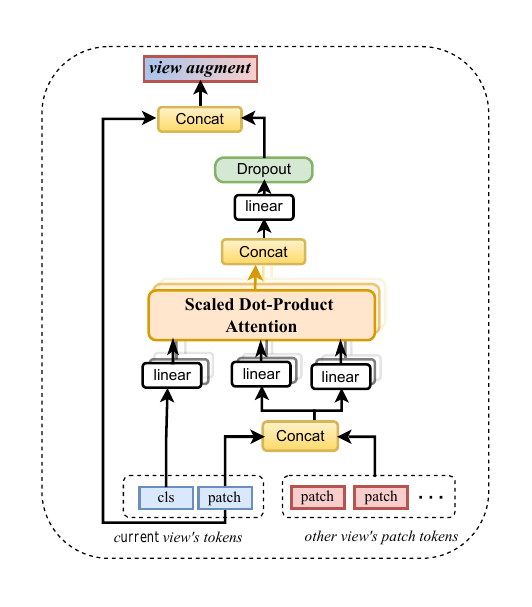}
    \vspace{-10mm}
    \caption{The view-specific feature enhancement sub-Module (VFEM). We perform cross-view calculations to extract enhanced view-specific features representations through cross-view feature correlation and transformation.
    }
    \label{fig:view_perception}
\end{figure}

% 全局语义引导,提取当前视角的CLS token作为全局语义指导，保留原始局部patch特征以维持视角特异性。多视角特征聚合,将其他视角的局部patch特征展开为序列
View-specific Feature Enhancement sub-Module. As show in Fig.~\ref{fig:view_perception}, we propose View-specific Feature Enhancement sub-Module to enhance these features using a cross-view attention mechanism. View-specific Feature Enhancement sub-Module captures the inter-dependencies between features from different viewpoints to enhance view-specific information. 
Global Semantic Guidance. Extract the cls token of the current view as the global semantic guidance ($x_{cls} \in \mathbb{R}^D$), while retaining the original local patch features ($x_{patch} \in \mathbb{R}^{N_p \times D}$) to maintain local characteristics.
Multi-View Feature Aggregation. Flatten the local patch features of other views into a sequence ($y_{patch} \in \mathbb{R}^{(v-1) \times N_p \times D}$) and compute the spatial dependencies between the global semantic guidance and multi-view local features using a cross-attention mechanism:
% 跨视角感知公式
\begin{equation}
        \text{Attention}(Q,K,V) = \text{Softmax}\left(\frac{QK^T}{\sqrt{d_k}}\right)V ,
\end{equation}
where $Q=x_{cls}$ and $K=V=\text{Concat}(x_{patch}, y_{patch})$. Feature Combine. Concatenate the enhanced global semantic features with the original local patch features to form the final feature representation. This design injects complementary information across views while preserving the original local details.

\begin{equation}
    \begin{aligned}
        & F_{vip}^i   = Attention(F_i) ,  \\                     
        & F_{a}^i     = Concat(F^i, F_{vip}^i), 
    \end{aligned}
\end{equation}
where $ F_{cip}\in \mathbb{R}^{B \times 1\times N \times D}$ and $F_{a}^i\in \mathbb{R}^{B \times (view+1) \times N \times D}$.

% 该模块通过端到端训练，能够自适应地学习不同视角间的语义关联，生成具有强判别性的车辆特征表示，为后续的多查询重识别任务提供鲁棒的特征基础。
Through end-to-end training, this module adaptively learns relationships across different views, generating highly discriminative vehicle view-specific feature representations, thereby providing a robust feature foundation for subsequent multi-query ReID vehicle tasks.

\subsection{Dynamic Multi-view Fusion sub-Module (DMFM)}
\label{method:mdf}
% 为缓解多视图车辆重识别中视角差异带来的问题，我们提出了一种基于跨视图注意力引导的专家混合机制（Mixture of Experts, MoE）的多视图动态融合模块（Multi-View Dynamic Fusion, mdf）。该模块以跨视角感知模块（CVP）生成的多视图增强特征为输入，通过动态融合生成统一的特征表示。以下详细介绍该模块的设计与实现。
To address the challenge of view discrepancies in multi-view vehicle ReID, we propose the Dynamic Multi-view Fusion sub-Module (DMFM), which dynamically fuses multi-view features using a cross-view attention-guided Mixture of Experts mechanism. The DMFM sub-module takes the view-enhanced features from the VFEM as input and generates a unified feature representation. Inspired by the Mixture of Experts~\cite{Jiang2024MixtralOE} approach, this module learns to dynamically adjust the weights of different tokens through a routing mechanism. By doing so, it effectively integrates information from multiple views, resulting in a fused feature representation that captures the diverse perspectives. The DMFM module consists of four key components: Cross View Relation Modeling, View Attention Mechanism, Expert Networks, Gating Network.

\noindent
\textbf{View Attention Mechanism}. A view attention network computes attention weights to dynamically aggregate features across views.
$$
A(x) = \text{Softmax}(MLP(x)).
$$
\noindent
% 跨视角关系建模
\textbf{View Relation Modeling}. A relation modeling network computes view-specific relation scores to capture the dependencies between different views.
$$
R(x) =FC(x).
$$
% 专家网络
\noindent
\textbf{Expert Networks}. A set of K expert networks are employed to learn diverse feature transformations. Each expert network is implemented as a multi-layer perceptron (MLP) with ReLU activation.
$$
\text{Expert}_k(x) = FFN_k(x), \quad k \in \{1, \dots, K\} .
$$
% 门控网络

\noindent
\textbf{Gating Network}: A gating network computes weights for dynamically selecting and combining the outputs of the expert networks.
$$
G(x, R(x)) = \text{Softmax}(MLP(CAT(x,R(x)))).
$$

% 基于注意力权重聚合多视角特征, 保留关键视角信息，抑制噪声干扰。
% 动态特征融合,加权融合专家输出，并引入残差连接,其中\alpha (初始值0.5) 为可学习的残差缩放因子。
% $$
%    F_{fusion} = \sum_{k=1}^{K} G_k(F_{agg},R(F)) \cdot {E_k(F_{agg})}+\alpha \cdot F_{agg} 
% $$
Dynamic feature fusion is achieved by weighted combination of expert outputs and the introduction of residual connections, where $\alpha$ is a learnable residual scaling factor. The fusion process can be expressed as:
\begin{equation}
F_{fusion} = \sum_{k=1}^{K} G(F_{aug},R(F_a)) \cdot {Expert_k(F_{aug})}+\alpha \cdot F_{aug},
\end{equation}
where $F_{aug}$ is defined as:
$$
 F_{aug}= \sum_{i=1}^{V+1} A(F_a^i) \cdot {F_a^i} ,
$$
where $G$ represents the expert weights generated by the gating network, $Expert_k$ denote the outputs of the expert networks, $F_{a}$ is the aggregated feature, and $F_i$ is the $i$-th view feature. This formula enables adaptive feature fusion and information retention through dynamic weighting and residual connections.

% 优势：
% MDF 模块通过动态调节特征权重，
The DMFM module dynamically adjusts feature weights to ensure that the most representative features receive more attention during the fusion process.
%  DMFM 模块的通过基于注意力引导的MOE的融合机制确保了特征融合的灵活性，确保最具代表性的特征在融合过程中获得更多的关注.使模型能够在不同的时间、视角下保持高精度的重识别能力。
The DMFM module employs an attention-guided MoE mechanism to achieve flexible and adaptive feature fusion, ensuring that the most representative features are prioritized during the integration process. This design enables the model to maintain high-precision ReID capabilities across diverse temporal and viewpoint conditions.

\subsection{Multi-view Alignment Loss (MAL)}
\label{method:mal}
% 为约束 DMFM 模块输出的融合特征与单视角特征的一致性，我们设计了多视角对齐损失（Multi-view Alignment Loss, LMAL。该损失由重建损失和对比损失两部分组成，分别从特征尺度和语义空间两个维度对融合特征进行约束，确保其既能保留单视角特征的细节信息，又能对齐多视角间的语义一致性。
To ensure that the fused features maintain consistency with the single image features from the DMFM module while preserving multi image semantic alignment, we propose the Multi-view Alignment Loss (MAL, $\mathcal{L}_{MAL}$), which combines contrastive loss and reconstruction loss to jointly constrain feature representations in both scale and semantic space. The contrastive loss minimizes feature discrepancies to retain single-view details, while the reconstruction loss enforces semantic coherence across different images, achieving a balanced fusion that captures fine-grained visual cues while maintaining robust cross-view consistency. Given the fused feature $F_{fusion} \in \mathbb{R}^{B \times ({N_p+1}) \times D}$ and the single images features $\{F_i\}_{i=1}^V \in \mathbb{R}^{B \times V \times ({N_p+1}) \times D}$, where $B$ is the batch size, $V$ is the number of views,$N_p$ is the number of patches, and $D$ is the feature dimension, the total loss $\mathcal{L}_{\text{MAL}}$ is defined as:
\begin{equation}
    \mathcal{L}_{MAL} = (1 - \alpha) \cdot \mathcal{L}_{contrast} + \alpha \cdot \mathcal{L}_{recon},
\end{equation}
where $\alpha \in (0, 1)$ is a weighting coefficient that balances the contributions of the two loss components.

\noindent
\textbf{Contrastive Loss.} The contrastive loss maximizes the semantic consistency between the fused feature and the single images features in the normalized feature space using bidirectional contrastive learning.
\begin{equation}
\mathcal{L}_{contrast} = \frac{1}{V} \sum_{i=1}^V \left( \mathcal{L}_{view}(F_{fusion}, F_i) + \mathcal{L}_{view}(F_i, F_{fusion}) \right),
\end{equation}
where the view loss $(\mathcal{L}_{view})$ is defined as:
$$
\mathcal{L}_{\text{view}}(x, y) = -\frac{1}{B} \sum_{j=1}^B \log \frac{\exp(x_j^\top y_j / \tau)}{\sum_{k=1}^B \exp(x_j^\top y_k / \tau)},
$$
where $\tau$ is the temperature coefficient that controls the sharpness of the similarity distribution.

\noindent
\textbf{Reconstruction Loss.} The reconstruction loss constrains the alignment between the fused feature and each single-view feature in the original feature scale using Mean Squared Error:

\begin{equation}
\mathcal{L}_{recon} = \frac{1}{V} \sum_{i=1}^V \|F_{fusion} - F_i\|_2^2 ,
\end{equation}
$\mathcal{L}_{recon}$ ensures that the fused feature retains the detailed information of the single-view features, avoiding excessive smoothing.

%MAL 损失函数介绍
\begin{figure}[]
    \includegraphics[width=1.0\columnwidth]{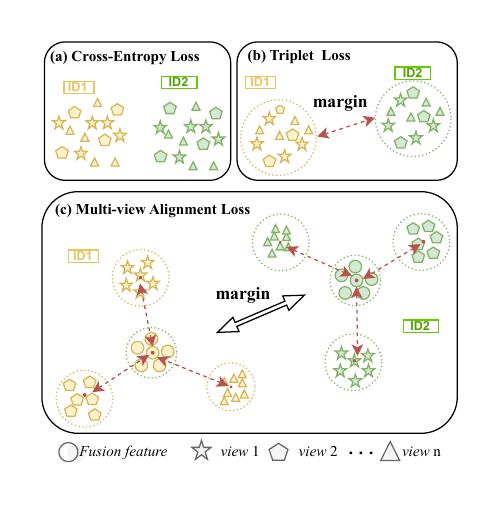}
    \vspace{-10mm}
    \caption{Multi-view Alignment Loss (MAL). MAL encourages the model to learn more comprehensive and balanced feature representations. This design is particularly suitable for tasks requiring the fusion of multi-source information, as it helps the model find the optimal balance between different views and capture richer feature information.}
    \label{fig:cvloss}
\end{figure}

\begin{equation}
\label{eq:ltotal}
\mathcal{L}_{total} =\mathcal{L}_{MAL}+\mathcal{L}_{reid} ,
\end{equation}

CAFNet is enhanced by the co-optimization of the VFEM and DMFM modules, where multi-query features are constrained under the \eqref{eq:ltotal} loss. VFEM extracts and enhances complementary features for each viewpoint using cross-attention, while DMFM dynamically evaluates feature importance through a gating network, enabling adaptive fusion for unified representation. Their synergy ensures effective perspective fusion and balanced feature contributions, improving overall performance.

\section{Dataset and Evaluation Protocols}
\label{sec:dataset}
\subsection{Description of LCRI-1K}
We introduce a large-scale dataset, named Large-scale Cross-camera ReID for Vehicles (LCRI-1K), which is captured from multiple cameras in real-world traffic environments.
\begin{table*}[t]
\caption{
Publicly available benchmark datasets for vehicle ReID.
Comparisons among VehicleID~\cite{liu2016deep_vehiclesID}, VeRI-776~\cite{liu2016deep_veri776}, CityFlow~\cite{tang2019cityflow}, VERI-Wild~\cite{liu2020beyond_vehicle-wild}, MuRI~\cite{zheng2023multi}, and the created LCRI-1K datasets for vehicle ReID.}
    \resizebox{\linewidth}{!}{
        \begin{tabular}{|c|c|c|c|c|c|c|}
        \hline
        Dataset            & VeRI-776\cite{liu2016deep_veri776}  & VehicleID~\cite{liu2016deep_vehiclesID} & VERI-Wild~\cite{liu2020beyond_vehicle-wild} & CityFlow~\cite{tang2019cityflow} & MuRI~\cite{zheng2023multi}  & LCRI-1K (ours)  \\ \hline
        Images             & 49,360   & 221,763   & 416,314      & 56,277    & 23,637   & \textbf{107,805}       \\ \hline
        Identities         & 776      & 26,267    & 40,671       & 666       & 200      & \textbf{1,090}          \\ \hline
        Cameras            & 18       & 40        & 174          & $-$         & 6,142    & \textbf{23,637}         \\ \hline
        Cross Cameras/id   & 4.2      & $-$       & 4.0          & about 2   & 5.0      & \textbf{67.5}           \\ \hline
        Views              & 6        & 2         & $Unconstrained$ &$-$      & $-$   & \textbf{$Unconstrained$}  \\ \hline
        Cross-resolution   & $\times$ & $\times$  & $\times$     & $-$       & $\surd$  & $\surd$          \\ \hline
        Occlusion          & $\times$ & $\times$  & $\surd$      & $\surd$   & $\surd$  & $\surd$          \\ \hline
        Complex Background & $\times$ & $\times$  & $\surd$      & $\surd$   & $\surd$  & $\surd$          \\ \hline
        Capture Time       & $18h$      & $-$       & a month    & $-$     & $-$      & \textbf{a year}  \\ \hline
        Morning            & $\times$ & $-$       & $\surd$      & $-$       & $\surd$  & $\surd$          \\ \hline
        Afternoon          & $\surd$  & $-$       & $\surd$      & $-$       & $\surd$  & $\surd$          \\ \hline
        Night              & $\times$ & $-$       & $\surd$      & $-$       & $\surd$  & $\surd$          \\ \hline
        \end{tabular}
    }
\label{tab:datacompar}
\end{table*}

\begin{figure}[]
\centering
    \includegraphics[width=0.9\columnwidth]{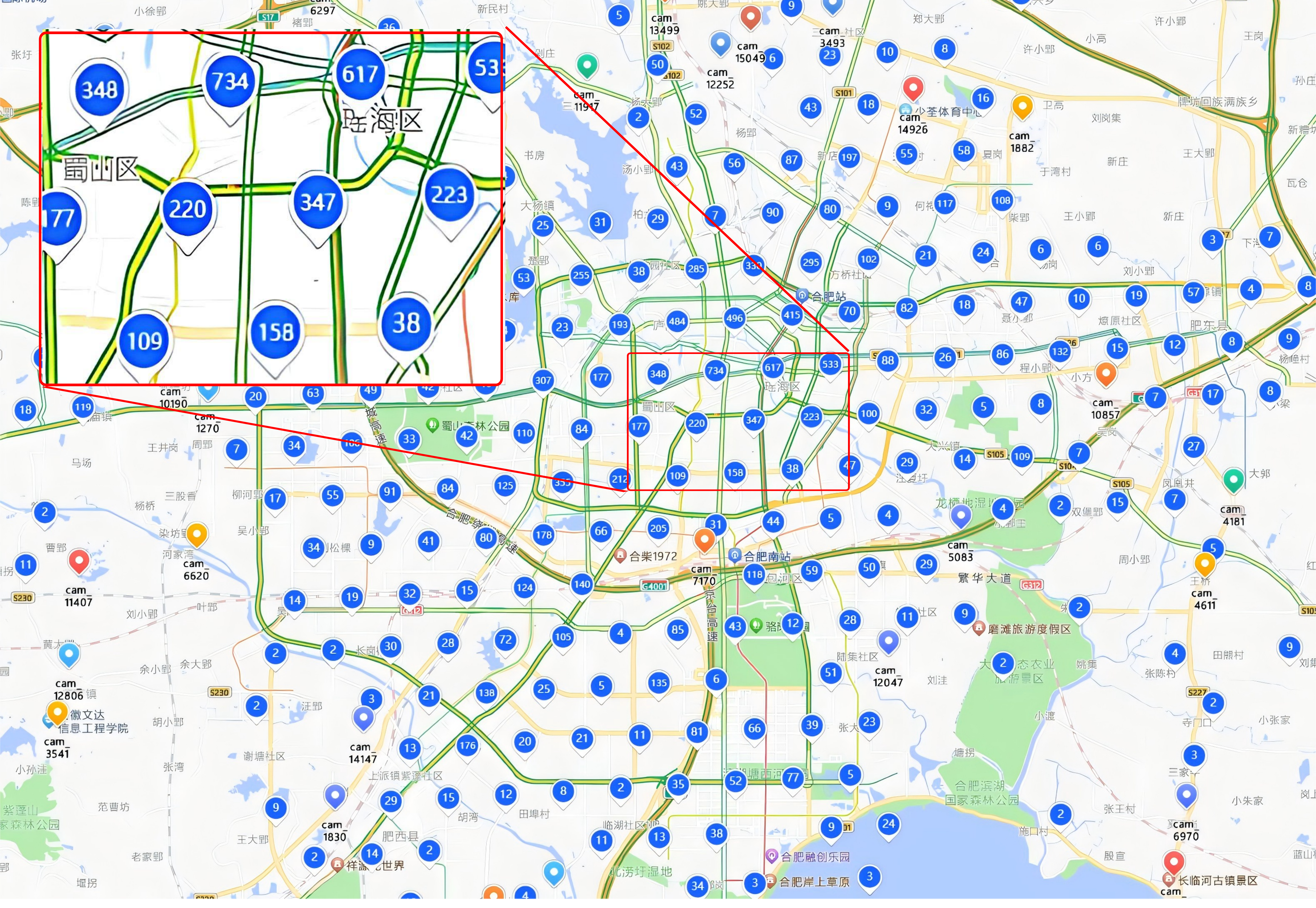}
    \vspace{2mm}
    \caption{
    Camera distribution in the LCRI-1K. The blue circles represent the approximate distribution of cameras on the map, and the numbers in the blue circles represent the number of cameras distributed in the area.}
    \label{fig:cam_dis}
\end{figure}

\begin{figure}[ht]
    \begin{center}
        \includegraphics[width=0.8\columnwidth]{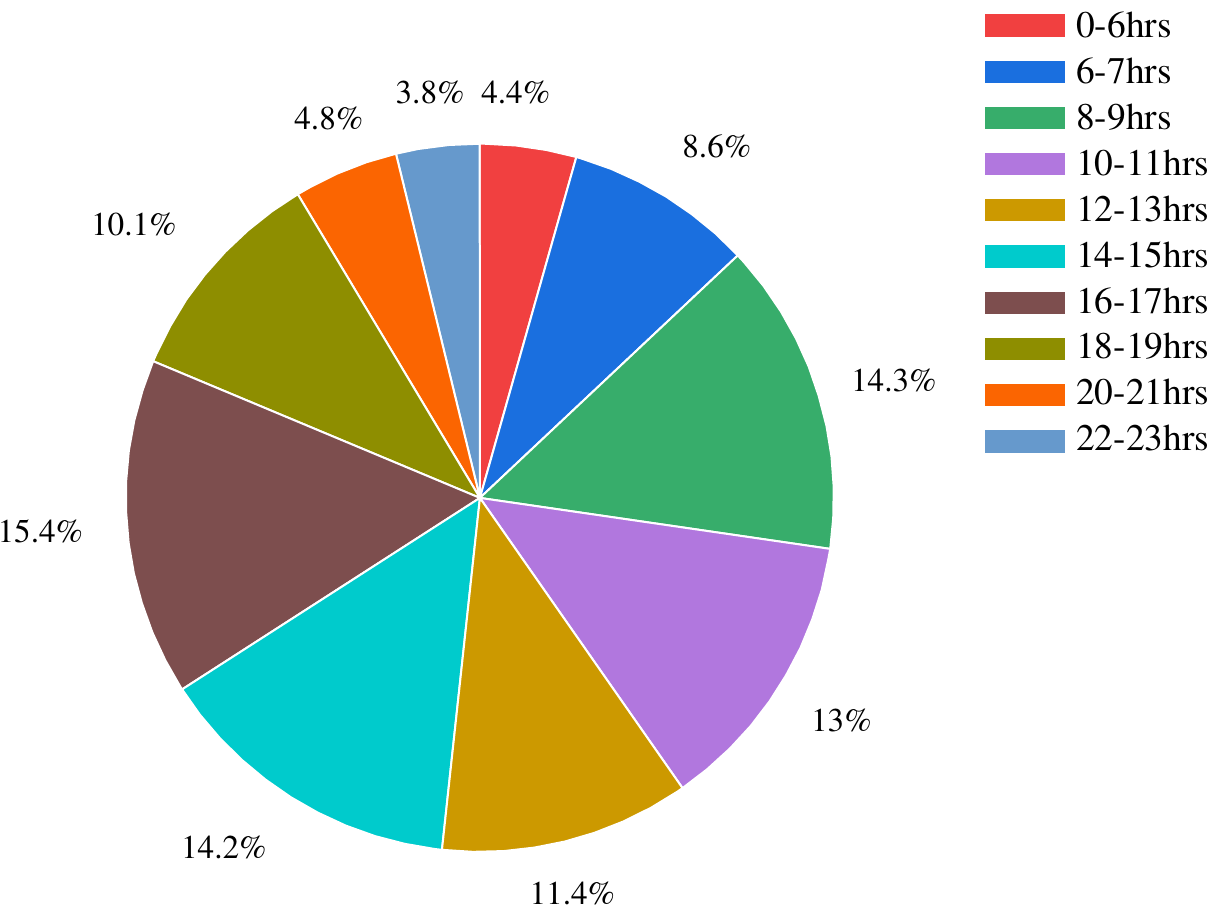}
    \end{center}
    \vspace{-4mm}
    \caption{
    The distribution of images across time. The distribution of images across time shows vehicle data from different times of day over a 24-hour period, with each time segment represented in a distinct color.}
    \label{fig:distime}
\end{figure}

The LCRI-1K dataset is a full range of vehicle data collected in a large urban traffic road system covering more than 1,339 square kilometers, covering all suitable vehicle samples. Fig.~\ref{fig:cam_dis} shows the camera distribution across urban traffic arteries, offering diverse real-world data while presenting challenges like lighting variations, occlusions, and viewpoint differences for vehicle ReID. We mask the vehicle's license plate information. As shown in Fig.~\ref{fig:distime}, in the spatial dimension, compared with existing vehicle datasets such as MURI~\cite{zheng2023multi}, VeRi776~\cite{liu2016deep_veri776}~\emph{etc.}, our dataset has a significant increase in the number of cross-cameras, and the scene diversity and challenges are richer. In terms of the time dimension, it covers the 24-hour traffic conditions for a whole year. Each vehicle image was labeled with a corresponding camera ID.

In the query set, we select 15 images for each vehicle, which were evenly distributed over a 24-hour period. To ensure comprehensive coverage of the data across the temporal dimension, we divide the day into multiple time periods and select image samples of vehicles from each time period. This strategy of uniform distribution ensures that the evaluation reflects the scenarios at different times throughout the day and enhances the effectiveness of the test across the temporal dimension. Meanwhile, all images are manually screened to avoid data redundancy, thus ensuring the accuracy and representativeness of the assessment. Aiming at the problems of strong spatio-temporal continuity and single perspective of query set samples caused by the common use of video frame extraction method in existing datasets, this study carefully constructs the query set through a manual screening mechanism by selecting 15 samples covering multiple perspectives and typical scenes for each ID, which effectively solves the limitations of the existing datasets, and significantly enhances the authenticity and challenge of cross-scene retrieval. The comprehensiveness and uniqueness of this dataset provides a more rigorous and reliable benchmark for cross-camera and cross-time generalization performance evaluation of vehicle ReID algorithms in complex urban environments, and provides important support for research in related fields.

% 数据集 id 跨相机展示
\begin{figure}[t]
  \centering
  \includegraphics[width=0.9\linewidth]{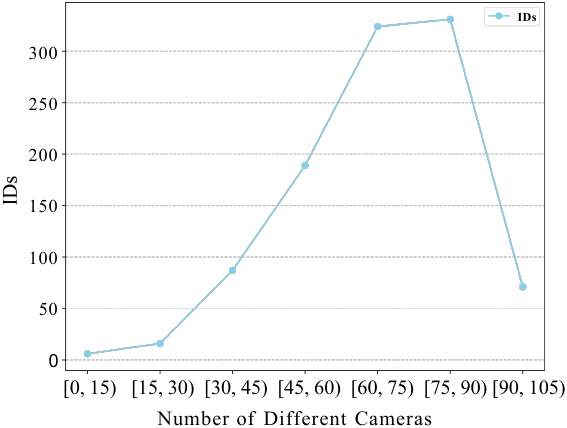}
  \vspace{-2mm}
  \caption{Statistics on the number of vehicles crossing the camera.}
  \label{fig:cam_cross_id}
\end{figure}

% 数据集相机分布展示

%图像的时间分布显示了全天 24 小时不同时间段的车辆数据，用不同颜色区分每个时间段。我们提出的数据集涵盖了完整的时间范围，确保车辆在各个时段均有出现。这一特性使得 LCRI-1K 数据集能够更全面地反映真实交通环境下的车辆再识别需求，提高模型在全天候场景中的适应性和鲁棒性。
% 24h 分布

% 数据集月份跨度展示
\begin{figure}[t]
  \centering
  \includegraphics[width=1.0\linewidth]{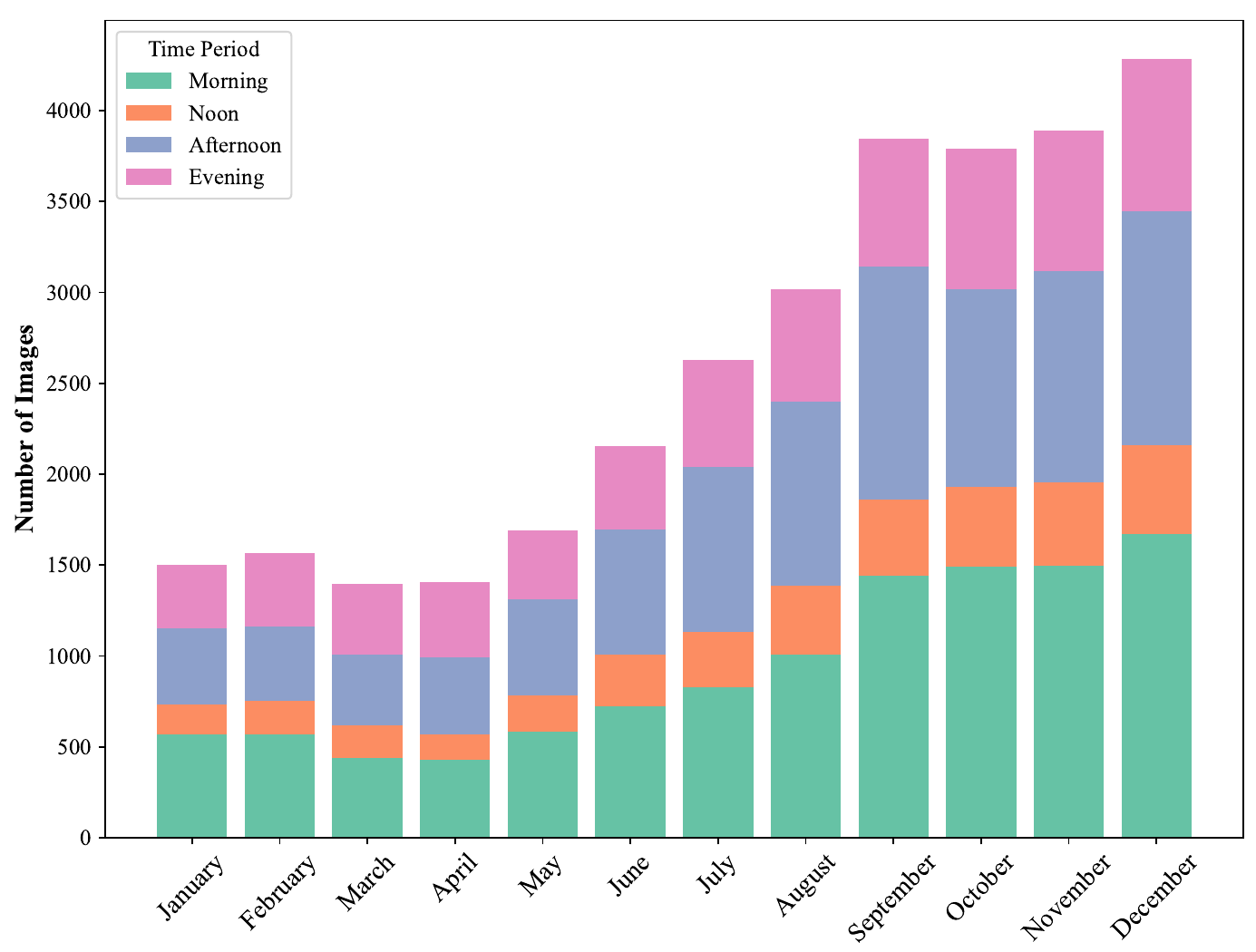}
  \caption{Image counts by month and time period. Spatiotemporal distribution of vehicle images across months and four daily periods (morning/noon/afternoon/night), demonstrating comprehensive temporal coverage.}
   \label{fig:mouth_cross}
\end{figure}

\noindent
\textbf{Dataset Challenges.}
Our LCRI-1K mainly contains five different challenges as shown in Table~\ref{tab:datacompar}.
First, our dataset provides comprehensive  viewpoints, including $front$ ($side$ $front$), $side$ and $rear$ ($side$ $rear$) for each vehicle, which produce huge intra-class for Vehicle ReID.
Then, due to the varying distances between cameras and vehicles of interest, the vehicle images present different resolutions as shown in Fig.~\ref{fig:challeng}. The poor detailed information in the low resolution, as well as the appearance gap between different resolutions, further bring huge challenges for vehicle ReID.

Moreover, as shown in Fig.~\ref{fig:cam_dis} LCRI-1K dataset is collected in a large city surveillance system spanned more than 11,445 $km^2$, and the urban environment is complex.
To this end, LCRI-1K contains many vehicle images with occluded and complex backgrounds, as shown in Fig.~\ref{fig:challeng} (a) and (b). This poses a serious challenge for vehicle ReID.
Fig.~\ref{fig:mouth_cross} shows that the vehicle images in the LCRI-1K dataset were collected over a long time span of more than one year, which provides a large amount of cross-temporal vehicle data under different lighting, with each vehicle appearing in a very comprehensive time span and spatial span, spanning a rich variety of scenarios, including morning, afternoon, and evening, as shown in Fig.~\ref{fig:challeng} (c).
Dramatic changes in illumination can lead to significant differences in vehicle appearance. In addition, the strong illumination of headlights and taillights, as well as street lamps at night, poses an additional challenge for vehicle ReID.

\noindent
\textbf{Dataset Characteristics.}
As shown in Fig.~\ref{fig:challeng} and Fig.~\ref{fig:cam_cross_id}, the dataset features the following key characteristics. Compared with existing prevalent ReID datasets, as shown in Table~\ref{tab:datacompar} and Table~\ref{tab:dataset_split} , LCRI-1K demonstrates several key advantages in terms of the average number of cross-camera appearances per ID. These advantages include:
\begin{enumerate}
    \item
    {\bf Numerous cameras with wide coverage}. LCRI-1K includes 1090 vehicle IDs captured by 23,637 cameras from a real-world transportation surveillance system, spanning over 11,445 km² of urban area.
    \item
    {\bf Comprehensive viewpoints and time of each ID}.
    LCRI provides comprehensive five viewpoints, including $front$ ($side$ $front$), $side$ and $rear$ ($side$ $rear$) for each vehicle, which provides a more realistic and challenging scenario for vehicle ReID.
    \item
    {\bf Large number of cameras crossed by each ID}.
    As show in Fig~\ref{fig:cam_dis}, each vehicle in LCRI-1K crosses 67.5 cameras in average, varying from 9 to 90 cameras. Based on viewpoint and background information, we construct multi-query task groups that closely mimic real traffic scenarios. These groups include diverse viewpoint combinations and complex background interferences, offering a high-difficulty benchmark for multi-query ReID research. The LCRI-1K dataset provides a solid foundation for multi-query vehicle ReID research, driving deeper exploration of the task and laying critical data support for future studies.
\end{enumerate}

\subsection{Evaluation Metric}
The LCRI-1K is manually divided into two parts for training and testing, as shown in Table~\ref{tab:dataset_split}. 
\begin{table}
\renewcommand\arraystretch{1.3}
\setlength\tabcolsep{5pt}
\normalsize
\centering
\caption{The splitting for training and testing sets. (IDs/Images)}
    \resizebox{\linewidth}{!}{
        \begin{tabular}{lcccc}
        \toprule
        \textbf{Dataset}                          & \textbf{Train} & \textbf{Query}  & \textbf{Gallery} \\
        \midrule
        VehicleID\cite{liu2016deep_vehiclesID}     & 13,164/100,182 & 2,400/2,400     & 2,400/17,638    \\
        VeRI-776\cite{liu2016deep_veri776}         & 576/37,778     & 200/1,678       & 200/11,579      \\
        VeRI-Wild\cite{liu2020beyond_vehicle-wild} & 30,671/277,797 & 10,000/10,000   & 10,000/128,517  \\
        MURI\cite{zheng2023multi}                  & 150/18308      & 50/150          & 50/5329         \\
        LCRI-1K                                    & 801/80,096     & 289/4335        & 289/24450       \\
        \bottomrule
        \end{tabular}
    }
\label{tab:dataset_split}
\end{table}
In the multi-query ReID process, for each given query group, a candidate list sorted by the feature distances between the query group and reference images is returned from the database. 
We adopt the Cumulative Matching Characteristics (CMC)~\cite{liu2016deep_veri776} with Rank-k matching accuracy, mean Average Precision (mAP)~\cite{zheng2015scalable}, mean Inverse Negative Penalty (mINP)~\cite{ye2021deep} as the evaluation metrics. The red and blue respectively represent the first and second results.

%%CMC
Cumulative Match Characteristics (CMC) curve shows the probability that a query identity appears in different-sized candidate lists. Mean Average Precision (mAP) evaluates the overall performance for ReID.

% The cumulative match characteristics at rank $k$ can be calculated as:
% \begin{equation} CMC@k = \frac{\sum_{q=1}^{Q} gt(q, k)}{Q} \end{equation}
% where gt(q,k) equals 1 when the ground truth of q image appears before rank $k$.

%%mAP

% and is defined as follows:
% \begin{equation}
% AP = \frac{\sum_{k=1}^{n} P(k) \times gt(k)}{N_{gt}}, mAP = \frac{\sum_{q=1}^{Q} AP(q)}{Q},
% \end{equation}
% where k is the rank in the recall list of size n, and $N_{gt}$ is the number of relevant vehicles. $P(k)$ is the precision at cutoff $k$ and $gt(k)$ 
% indicates whether the k-th recall is correct or not. $Q$ is the number of total query images. Moreover, Top $K$ match rate is also reported in the experiments.

%%minp
Inverse Negative Penalty (INP) evaluates the ability to retrieve the hardest correct match, providing a supplement for measuring the ReID performance, and is defined as follows:

\begin{equation}
    \begin{aligned}
    NP_i &= \frac{R_i^{\text{hard}} - |G_i|}{R_i^{\text{hard}}},\\
    mINP &= \frac{1}{n} \sum_i (1 - \text{NP}_i),
    \end{aligned}
\end{equation}
where $R^{hard}_i$ indicates the rank position of the hardest match, and $G_i$ represents the total number of correct matches for query $i$.
Mean Cross Scene Precision (mCSP) enhances mAP by penalizing positive samples with similar viewpoints from the same camera, offering a more realistic measure for ReID applications, and is defined as follows:
%% mCSP
\begin{equation}
     mCSP = \frac{\sum_{i=0}^{N_{cs}} \frac{TP - SC}{TP - SC + FP}}{N_{cs}},
\end{equation}
where $SC$ to denote the number of samples with the similar viewpoint under the same camera ID in $TP$, and $FP$ denotes the number of positive samples with prediction errors,and $N_{cs}$ denotes the captured target images from positive samples with different cameras.

\section{Experiments}
\label{sec:experiments}

\begin{table*}[!ht] 
\caption{Performance comparisons on MURI and LCRI-1K benchmark. Single represents the traditional single-query approach. score\_avg denotes averaging the similarity scores in the inference phase; feat\_avg denotes averaging the features and then calculating the similarity to obtain the final matching result; and multi-query denotes learning a unified feature representation.}
    \resizebox{\linewidth}{!}{
        \begin{tabular}{ccl|cccc|cccc}
            \hline
            \multirow{2}{*}{method}    & \multirow{2}{*}{Venue} & \multirow{2}{*}{\begin{tabular}[c]{@{}l@{}}infer\\ method\end{tabular}} & \multicolumn{4}{c|}{MURI~\cite{zheng2023multi}} & \multicolumn{4}{c}{LCRI-1K}   \\ \cline{4-11} 
                                       &                     &                        & mAP  & mCSP & mINP  & Rank1   & mAP   & mCSP  & mINP    & Rank1 \\ \hline
            \multirow{3}{*}{DMML~\cite{chen2019deep}}        &        & single        & 36.2 & 19.8 & 6.0   & 64.4    & 21.2  & 37.8  & 0.9    & 64.7 \\
                                       & {ICCV'19}                    & score\_avg    & 32.2 & 20.2 & 8.1   & 58.0    & 34.9  & 42.8  & 1.1    & 72.3 \\
                                       &                              & feat\_avg     & 52.4 & 27.5 & 8.8   & 76.6    & 44.8  & 39.5  & 3.5    & 74.9 \\ \hline
																																			    	  
            \multirow{3}{*}{RECT\_Net~\cite{zhu20rectnet}}   &        & single        & 41.5 & 22.5 & 7.4   & 72.4    & 44.8  & 4.0   & 1.2    & 80.4 \\
                                       &{CVPR'20}                     & score\_avg    & 48.6 & 35.5 & 7.6   & 82.0    & 60.4  & 64.4  & 2.3    & 90.1 \\
                                       &                              & feat\_avg     & 60.2 & 33.0 & 10.8  & 80.6    & 61.9  & 59.6  & 6.7    & \underline{91.6} \\ \hline
																																	   		    	  
            \multirow{3}{*}{TransReid~\cite{he2021transreid}} &       & single        & 61.2 & 42.0 & 9.8   & 72.5    & 56.4  & 57.8  & 2.8    & 84.3  \\
                                       & {ICCV'21}                    & score\_avg    & 75.6 & 69.2 & 17.9  & 95.2    & 64.5  & 64.5  & 8.3    & 89.6  \\
                                       &                              & feat\_avg     & 78.1 & 71.3 & \underline{29.1}&98.7   &\underline{66.2}  & 65.4  & \underline{14.0}   & 91.1  \\ \hline
																																			   
            \multirow{3}{*}{CAL~\cite{rao2021counterfactual}}&        & single        & 44.1 & 26.6 & 8.2   & 75.4    & 32.7  & 32.6  & 1.5    & 70.5  \\
                                       &{ICCV'21}                     & score\_avg    & 49.6 & 37.0 & 7.6   & 84.0    & 45.2  & 44.5  & 2.9    & 84.7  \\
                                       &                              & feat\_avg     & 64.5 & 35.9 & 13.8  & 81.6    & 48.5  & 46.0  & 3.2    & 84.1  \\ \hline
                                       
            \multirow{3}{*}{DRL-Net~\cite{he2021transreid}} &         & single        & 43.1 & 28.6 & 9.2   & 74.4    & 32.6  & 33.5  & 4.5    & 69.3  \\
                                       & {TMM'22}                     & score\_avg    & 60.6 & 57.5 & 10.6  & 85.0    & 46.2  & 45.5  & 5.9    & 85.7  \\
                                       &                              & feat\_avg     & 65.5 & 55.9 & 13.8  & 88.6    & 48.6  & 46.7  & 6.2    & 84.1  \\ \hline																													   
																																			    	  
            \multirow{3}{*}{PHA~\cite{zhang2023pha}}&                 & single        & 49.1 & 42.2 & 15.9  & 64.0    & 43.5  & 31.3  & 4.5    & 72.2  \\
                                       & {CVPR'23}                    & score\_avg    & 77.8 & 66.6 & 22.7  & 98.0     & 59.9  & 44.5  & 6.0    & 88.1  \\
                                       &                              & feat\_avg     & 76.6 & 65.4 & 23.6  & \underline{100.0}     & 60.1  & 46.1  & 6.0    & 88.1  \\ \hline

            \multirow{3}{*}{Fast REID  ~\cite{he2023fastreid}}&       & single        & 47.0 & 45.9 & 7.8   & 91.0    & 44.8  & 32.1  & 4.1    & 83.5 \\
                                       & {ACM MM'23}                  & score\_avg    & 57.3 & 50.3 & 12.7  & 92.1    & 58.7  & 37.5  & 6.2    & 87.8 \\
                                       &                              & feat\_avg     & 57.0 & 50.0 & 13.7  & 92.1    & 58.8  & 37.5  & 6.3    & 86.8 \\ \hline
																																			   
            \multirow{3}{*}{MVIIP    ~\cite{dong2024multi}}&          & single        & 47.0 & 45.9 & 7.8   & 90.0    & 44.7  & 32.0  & 4.1    & 83.8 \\
                                       & {INFFUS'24}                  & score\_avg    & 57.3 & 50.3 & 10.7  & 92.0    & 58.7  & 37.4  & 6.1    & 87.8 \\
                                       &                              & feat\_avg     & 57.0 & 50.0 & 10.7  & 92.0    & 58.7  & 37.4  & 6.2    & 87.8 \\ \hline
																																			   
            % \multirow{3}{*}{Fusion Reid~\cite{rao2021c}}&           & single        & 56.2 & 51.3 & 20.1  & 74.0    & 57.6  & 57.5  & 13.0   & 87.7  \\
            %                            & {TITS'25}                  & score\_avg    & 80.3 & 73.9 & 30.5  & 100.0   & 68.7  & 66.1  & 12.9   & 94.9 \\
            %                            &                            & feat\_avg     & 80.3 & 75.9 & 31.5  & 100.0   & 69.1  & 67.9  & 12.5   & 91.7 \\ \hline
		
           \multirow{3}{*}{RSFAN~\cite{xiong2025region}}&           & single          & 56.2  & 48.8  & 18.6  & 78.0   & 55.1  & 55.2  & 9.3    & 85.0   \\
                                       & {EAAI'25}                  & score\_avg      & 79.6  & 74.1  & 29.1  & 96.0   & 65.1  & 64.4  & 11.0   & 89.6  \\
                                       &                            & feat\_avg       &\underline{79.6}  &\underline{74.8}  & 29.1  & 96.0   &65.8  &\underline{65.4}  & 12.3   & 89.1  \\ \hline    
            VCNet~\cite{zheng2023multi}    &{TIP'23}                & multi-query     & 67.7 & 39.3 & 14.5  & 84.3     & 59.4  & 59.4  & 6.4    & 86.9 \\ \hline
            TOP REID~\cite{wang2024top}    &{CVPR'24}               & multi-query     & 75.2 & 73.6 & 19.5  & 100.0    & 58.9  & 60.2  & 8.4    & 88.3 \\ \hline
            CAFNet                         &{Ours}                  & multi-query    & \textbf{82.0} & \textbf{77.3} & \textbf{31.5}   & \textbf{100.0}  & \textbf{69.2}  & \textbf{66.5} & \textbf{17.6}    & \textbf{92.3} \\
            % CAFNet(+resnet)  & & multi query  & \textbf{83.0} & \textbf{76.1} & \textbf{33.1}   & \textbf{100.0}  & \textbf{68.9}  & \textbf{67.3} & \textbf{16.2}    & \textbf{90.6} \\
            \hline
        \end{tabular}
    }
\label{tab:sota1}
\end{table*}

\subsection{Implementation Details}
\noindent
{\bf Train.} We use ViT~\cite{alexey2020image} pre-trained on ImageNet-1k as our backbone. The model is trained for 180 epochs with the SGD optimizer.
We warm up the learning rate to 5e-2 in the first 5 epochs and the backbone is frozen in the warm-up step. 
The learning rate of 5e-2 is kept until the 60th, drops to 5e-3 in the 60th epoch, and drops to 5e-4 in the 75th epoch for faster convergence. 
We first pad 10 pixels on the image border, and then randomly crop it to 256×256. We also augment the data with random erasing. 
Further, we add a Batch Normalization layer after the global feature. A fully connected layer is added to map the global feature to the ID classification score. The batch size is 120 in the LCRI-1K dataset.

% 不同推理模式
% (a) 表示传统的单查询方法。(b)、(c)、(d) 分别代表多查询推理过程中的不同策略：(b) 表示在推理阶段取相似性得分的平均值；(c) 表示对特征进行平均后再计算相似性，以获得最终匹配结果；(d) 表示我们提出的方法，用于生成统一的特征表示。

% \begin{figure}[t]
%     \centering
%     \includegraphics[width=1\columnwidth]{img/inference.pdf}
%     \vspace{-4mm}
%     \caption{\textcolor{red}{Inference settings. (a) represents the single-query approaches. For multi-query inference scenarios, (b)(c) adopt similarity averaging and feature averaging strategies respectively, while (d) represents our proposed unified feature representation learning framework. AVG: averaging operation; colored blocks: distinct image features.}}
% \label{fig:inference}
% \end{figure}
\noindent
{\bf Inference.} In our inference process, we evaluate the methods in four inference ways, including single-query, score-average query ,feat-average query and multi-query. 
Single inference directly calculate the cosine distance between each query and the gallery set, which ignores the multi-view information during the inference. When faced with multiple queries, the intuitive inference way is the score average or feature average inference, which computes the average value of multiple query features or score. However, simply averaging the query features can not effectively use the different viewpoint information of the vehicle. To adaptively utilize the complementary information in the multiple queries from different cameras with diverse viewpoints combinations, we propose the multi-inference for the proposed CAFNet.
%对于查询集的图片，我们通过 CAFNet 方法获得融合后的特征；而对于图库集的每张图片，则直接通过 ViT backbone 提取特征
For the images in the query set, we obtain the fused features using the CAFNet method,while for each image in the gallery set, we directly extract features using the Vision Transformer(ViT)~\cite{alexey2020image} backbone.

% 对比实验
\subsection{Comparison with State-of-the-Art Methods}
To verify the effectiveness of the proposed CAFNet with the multi-query setting, we compare five state-of-the-art vehicle ReID methods on the collected LCRI-1K dataset and the MURI dataset. 
Specifically, we construct the multi-query setting with the number of queries $N_Q$ = 3.
We evaluated the state-of-the-art methods in both single query and score average query and feature average inferences for comparison. 
As shown in Table~\ref{tab:sota1}, 
%所有最先进的方法在平均推理方面相较于单一推理取得了显著进展，这表明使用多个查询可以更好地整合不同图像之间的互补信息。具体而言，相较于将相似度得分进行均值化，直接对特征进行均值化处理能取得更好的效果。虽然对特征进行均值化是一种简单的融合方法，但这一结果进一步证明了不同查询图像之间特征的互补性。
% 选取典型多模态行人重识别方法（TopReID），通过均值训练将其改进为多查询方法。实验表明，多模态领域的特征融合并不适合提升现有多查询方法的性能。
% 同时通过利用跨视角感知模块（VFEM）和通过视角动态融合模块（DMFM）有效整合来自不同视角的互补信息，我们的多查询CAFNet在车辆重识别任务中显著提升了性能。这种方法不仅捕捉了每个视角的独特特征，还通过动态调整不同特征的权重，增强了整体特征表示的能力。与现有最先进的技术相比，我们的方法通过提供更强的鲁棒性和辨识力，尤其在面对视角变化、光照条件和遮挡等挑战时，表现出了更优的性能。
all state-of-the-art methods have shown significant improvement in average inference compared to single-query inference, which demonstrates that using multiple queries can better integrate the complementary information between images. Specifically, averaging features yields better results than averaging similarity scores. Although averaging features is a simple fusion approach, this further proves the complementarity between features from different queries.
Moreover, we adopt TopReID~\cite{wang2024top}, a classic multi-modal ReID method, and optimize it into a multi-query model via mean training. Experiments show feature fusion in multi-modal domains fails to improve existing multi-query methods.
By leveraging the VFEM module and effectively integrating complementary information across different viewpoints through the Dynamic Multi-view Fusion sub-Module (DMFM), our CAFNet with multi-query significantly improves the performance of vehicle ReID. This approach not only captures the unique characteristics of each viewpoint, but also enhances the overall feature representation by dynamically adjusting the weights of different features. As a result, our method outperforms state-of-the-art techniques by providing more robust and discriminative feature representations, especially in challenging scenarios with varying viewpoints, lighting conditions, and occlusions.
% 实验结果表明，我们提出的方法在两个数据集上均实现了最先进的性能，特别是在mINP指标（反映最难样本的识别能力）上取得了显著提升。在Rank-1准确率和mAP（平均精度均值）方面，相较于第二名方法，我们的方法也表现出了明显的提升。这表明我们的模型不仅在整体性能上超越了现有的技术，同时在处理难度较大的样本时，展现了更强的鲁棒性和更高的准确性。
%
Experimental results demonstrate that our proposed method achieves state-of-the-art performance on both datasets, with a significant improvement in the mINP metric, which reflects the ability to handle the most challenging samples. In terms of Rank-1 accuracy and mAP, our method shows a clear improvement over the second-best approach. This indicates that our model not only outperforms existing methods in overall performance but also exhibits stronger robustness and higher accuracy when dealing with challenging samples.
This validates the effectiveness of the proposed CAFNet while handling the multi-query inference for vehicle ReID.

%% 总体消融实验
%为了验证各个模块对CAFNet-Network性能的贡献，我们在我们提出的数据集LCRI进行了消融实验。我们首先定义了一个基线模型，然后逐步增加我们所设计的模型包含所有模块（MVA和DMFM）。
% 我们采用了vit作为骨干网络来提取特征
%在表中，我们列出了不同模型配置的实验结果。例如，在增加多视角动态融合人模块（DMFM）后，模型的mAP从56.8%增加到67.6%，说明DMFMM在整合不同视角特征的重要作用，此外，增加跨视角感知（VFEM）后，模型性能进一步增加到68.8%，这表明VFEM在捕捉多视角特征依赖性方面的重要性。

\subsection{Ablation Studies}
To verify the contribution of individual modules to the performance of CAFNet, we conducted ablation experiments on our proposed dataset LCRI-1K. We first define a baseline model and then gradually increased our designed model to include all modules (i.e., VFEM, DMFM and MAL ) as shown in Table~\ref{tab:ablation} and Table~\ref{tab:ablation_loss}.
We employ Vision Transformer(ViT)~\cite{alexey2020image} in a single query fashion as our baseline to extract vehicle features.
Table~\ref{tab:ablation} present the experimental results of different model configurations. For example, after adding the Dynamic Multi-view Fusion sub-Module (DMFM), the mAP of the model increased from 65.2\% to 67.6\%, indicating the important role of DMFM in integrating features from different perspectives. In addition, after adding View-specific Feature Enhancement sub-Module (VFEM), the model performance further increased to 68.0\%, which shows the importance of VFEM. EV-MoE consistently brings a significant improvement on all the metrics by adaptively fusing the generated viewpoint weights with appearance features.

% 总体消融
\begin{table}
    \centering 
    \caption{Evaluation of VFEM, DMFM module under the number of queries $N_Q=3$}
    \resizebox{1.0\linewidth}{!}{
        \begin{tabular}{l|cccc}
            \hline
            Settings           & \multicolumn{1}{c}{mAP}  & \multicolumn{1}{c}{mCSP} & \multicolumn{1}{c}{mINP} & \multicolumn{1}{c}{Rank1}  \\ \hline
            (a) Baseline + Score average        & 62.6   & 61.0   & 12.2  & 91.8    \\
            (b) Baseline + Feat average         & 65.2   & 61.3   & 13.5  & 91.4    \\ 
            (c) Baseline + DMFM                 & 67.6   & 65.8   & 15.6  & 92.3    \\ 
            (d) Baseline + DMFM + VFEM          & 69.2   & 66.5   & 17.6  & 92.3    \\ \hline
        \end{tabular}
    }
    \label{tab:ablation}
\end{table}

%% VFE消融实验
\begin{table}
\centering
\caption{Evaluation of View-specific Features Enhancement sub-Module (VFEM) when the number of queries $N_Q=3$}
    \resizebox{1.0\linewidth}{!}{
        \begin{tabular}{l|cccc}
            \hline
            Settings          & \multicolumn{1}{c}{mAP}  & \multicolumn{1}{c}{mCSP} & \multicolumn{1}{c}{mINP} & \multicolumn{1}{c}{Rank1}  \\ \hline
            (a) Add                            & 58.3   & 55.0   & 7.2   & 83.2      \\
            (b) TPM\cite{wang2024top}          & 62.3   & 61.0   & 12.1  & 89.1      \\
            (c) CA\cite{mao2023cross}          & 67.1   & 65.3   & 15.8  & 91.6      \\
            (d) VFEM (ours)   & \textbf{69.2}   & \textbf{66.5}   & \textbf{17.6}  &\textbf{92.3}      \\ \hline
        \end{tabular}
    }
    \label{tab:ablation_cvp}
\end{table}

\noindent
\textbf{Evaluation on VFEM.}
%表[X]对比了四种跨视角特征融合方法在3查询条件下的性能表现。其中简单特征相加方法（Add）作为基线仅获得58.3%的mAP，基于多模态ReID的跨注意力方法（TPM）由于在多查询任务中存在模态冲突问题，其mINP指标仅为12.1%。标准交叉注意力机制（CA）虽然取得了65.3%的mCSP，但未能显式建模视角差异。我们提出的VFEM方法通过解耦视角不变特征和视角特异特征，最终以69.9%的mAP和92.3%的Rank-1准确率全面领先，其中mINP指标达到18.6%，较最优的CA方法提升2.8%，这验证了在多查询场景下显式建模视角交互的有效性。特别值得注意的是，VFEM在mINP指标上的优势比Rank-1更为显著，表明该方法对困难样本的识别能力具有更明显的提升效果。
Table~\ref{tab:ablation_cvp} compares the performance of four cross view feature fusion methods under query number $N_Q=3$ condition. The baseline method Add (feature addition) achieves 58.3\% mAP, while the TPM~\cite{he2021transreid} method based on multi-modal ReID shows limited performance (mINP 12.1\%) due to modality conflicts. The standard cross-attention CA~\cite{chen2021crossvit} achieves 65.3\% mCSP~\cite{zheng2023multi} but fails to explicitly model view differences. Our proposed VFEM method, which decouples view-specific features, demonstrates comprehensive superiority with 69.2\% mAP and 92.3\% Rank-1 accuracy. Notably, it achieves 17.6\% mINP (1.8\% improvement over CA~\cite{chen2021crossvit}), significantly enhancing the recognition capability for hard samples.

%% DMFM 消融实验
\begin{table}[]
\centering 
\caption{Evaluation of Dynamic Multi-view Fusion sub-Module (DMFM) when the number of queries $N_Q=3$}
    \resizebox{1.0\linewidth}{!}{
        \begin{tabular}{l|cccc}
            \hline
            Settings                     & \multicolumn{1}{c}{mAP}  & \multicolumn{1}{c}{mCSP} & \multicolumn{1}{c}{mINP} & \multicolumn{1}{c}{Rank1}  \\ \hline
            (a)Average                  & 65.3   & 64.2   & 12.1  & 88.1     \\
            (b)VAF~\cite{zheng2023multi} & 63.2   & 59.7   & 9.0   & 83.2     \\
            (c)MLP                      & 66.3   & 65.1   & 12.3  & 90.2     \\
            (d)DMFM (ours)              & \textbf{69.2}   & \textbf{66.5}   & \textbf{17.6}  & \textbf{92.3}     \\ \hline
        \end{tabular}
    }
    \label{tab:ablation_mdf}
\end{table}

\noindent
\textbf{Evaluation on DMFM.}
To further demonstrate the effectiveness and applicability of Dynamic Multi-view Fusion sub-Module (DMFM), We compare the existing methods based on viewpoints feature fusion VAF~\cite{zheng2023multi} module and 
 a MLP module. In vehicle ReID, the main challenge is the intra-class variability and inter-class similarity problem due to the difference in vehicle viewpoints. We can use the images from different views of the vehicle during the inference through multi-query. DMFM can assign weights adaptively according to the attention score between the viewpoints of vehicles in query. As shown in Table~\ref{tab:ablation_mdf}, after integrating the proposed DMFM into Vision Transformer(ViT)~\cite{alexey2020image}, it brings a large margin improvement over the original methods by fusing the input feature and view augmentation features from different view. This verifies that DMFM can better integrate the complementary information among different viewpoints.

\noindent
\textbf{Evaluation on MAL.}
The ablation study (Table~\ref{tab:ablation_loss}) shows our Multi-view Alignment Loss (MAL, $\mathcal{L}_{MAL}$) boosts multi-query ReID performance when combined with standard ID and triplet losses. The baseline (ID+triplet) achieves 65.2\% mAP and 87.3\% Rank-1, while adding MAL alone improves these to 68.3\% (+3.1\%) and 91.4\% (+4.1\%), with mINP increasing by 2.6\%. The full combination reaches 69.2\% mAP (+3.9\%) and 92.3\% Rank-1 (+5.0\%), with mINP showing the largest gain (+5.1\%), confirming MAL's effectiveness for hard cases and its synergy with conventional losses.

%损失消融 
\begin{table}[t]
\renewcommand\arraystretch{1.0}
\setlength\tabcolsep{5pt}
\normalsize
\centering 
\caption{Evaluation of Multi-view Alignment Loss (MAL) when the number of queries $N_Q=3$}
    \begin{tabular}{l|cccc}
        \hline
        Settings            & \multicolumn{1}{c}{mAP} & \multicolumn{1}{c}{mCSP} & \multicolumn{1}{c}{mINP} & \multicolumn{1}{c}{Rank1} \\ \hline
        (a) $\mathcal{L}_{ID}+\mathcal{L}_{triplet}$           & 65.2   & 61.3  & 13.5  & 87.3    \\
        (b) $\mathcal{L}_{MAL}$                                & 68.3   & 66.4  & 16.1  & 91.4    \\
        (c) $\mathcal{L}_{MAL}+\mathcal{L}_{ID}+\mathcal{L}_{triplet}$  & \textbf{69.2}   & \textbf{66.5}  & \textbf{17.6}  & \textbf{92.3}    \\ \hline
    \end{tabular}

    \label{tab:ablation_loss}
\end{table}

% 不同查询数量实验
% 我们还测试了不同查询数量（2到10张）下的模型性能变化，结果显示当查询图片数量达到3到4张时，模型性能提升最快；随后，随着查询图片数量的进一步增加，模型性能的提升幅度减小。这表明3到4张图片已能够完整表达车辆的整体特征，而更多的查询图片则增加了信息冗余。此外，随着查询图片数量的增加，低质量的图片（如遮挡、暗光、运动模糊等）可能掺入其中，对有效特征信息的贡献较小甚至干扰。然而，尽管如此，模型性能仍然随查询图片数量增加而提升，说明我们的网络不仅能够有效融合多张图片的信息，还对噪声具有很强的鲁棒性。
\begin{table}[t]
\renewcommand\arraystretch{1.0}
\setlength\tabcolsep{5pt}
\centering
\normalsize
\caption{% 在CAFNet上使用不同的查询数量(2~10)测试
    Evaluation CAFNet with different query number on LCRI-1K.}
    \begin{tabular}{l|cccc}
        \hline
        Query number    & mAP    & mCSP  & mINP & Rank1 \\ \hline
        (a) 2 query     & 65.7   & 64.0  & 7.6 & 89.3 \\ 
        (b) 3 query     & 69.2   & 66.5  & 17.6  & 92.3 \\ 
        (c) 4 query     & 70.8   & 68.3  & 18.5  & 93.7 \\ 
        (d) 5 query     & 71.0   & 69.2  & 18.6  & 94.1 \\ 
        (e) 6 query     & 71.3   & 69.4  & 18.5  & 95.7 \\ 
        (f) 7 query     & 72.1   & 70.3  & 18.6  & 96.9 \\ 
        (g) 8 query     & 72.5   & 71.0  & 18.8  & 96.9 \\ 
        (h) 9 query     & 72.8   & 71.3  & 18.8  & 96.3 \\
        (i) 10 query    & \textbf{73.1}   & \textbf{71.1}  & \textbf{18.8}  & \textbf{96.3} \\\hline
    \end{tabular}
    \label{tab:query_num}
\end{table}

\subsection{Evaluation on query numbers}
As shown in Table~\ref{tab:query_num}, we also tested the model's performance with varying numbers of query images (ranging from 2 to 10). The results show that performance improves most significantly when the number of query images is between 3 and 4. After this point, the performance gains diminish as the number of query images increases. This suggests that 3 to 4 images are sufficient to fully capture the vehicle's overall features, while additional images mostly add redundant information. Furthermore, as the number of query images increases, low-quality images (such as those affected by occlusion, low light, or motion blur) may be included, which contribute less to the effective feature information and may even introduce noise. Nevertheless, the model's performance continues to improve as the number of query images increases, indicating that our network is not only effective in integrating information from multiple images but also exhibits strong robustness to noise.

% 随机掩码实验
% 真实场景下车辆容易受其他车辆遮挡导致查询精度降低，这种情况下前景车辆的遮挡常常会误导模型提取目标车辆的特征，从而误导判断.基于这种情况，本文提出的车辆多查询网络通过多个视角多张图片特征的互补和融合能够有效降低前景车辆的干扰。为了模拟这种情况我们设计了通过对原始车辆图片添加随机概率的类似于cutmix方式,使用非目标车辆的生产掩码完全遮挡目标车辆来模拟真实复杂的交通环境中车辆拥堵导致的车辆遮挡车辆，实验结果如~\ref{fig:randmask}

\begin{figure}
  \centering
  \includegraphics[width=1.0\linewidth]{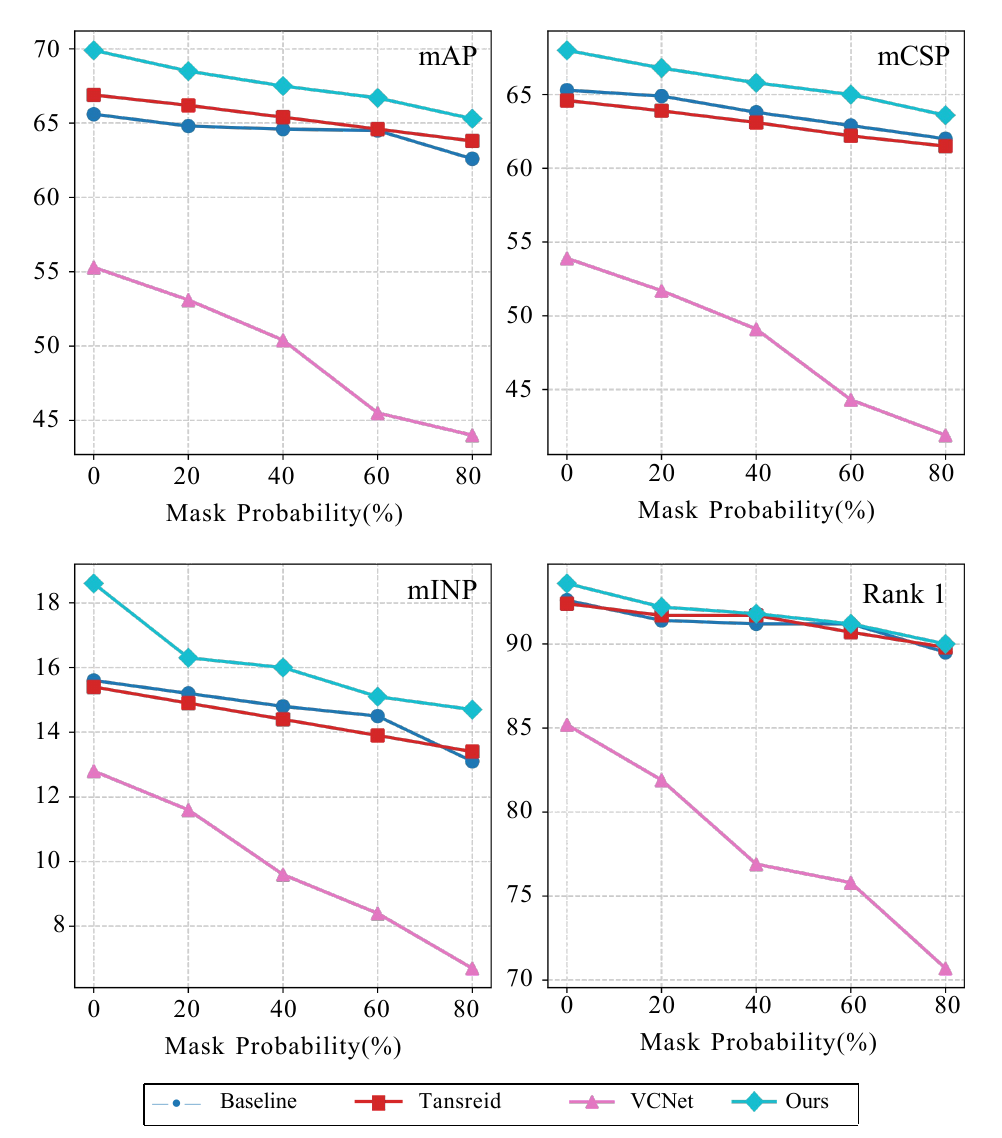}
  \vspace{-4mm}
  \caption{Random mask Occlusion.We designed five groups with varying occlusion probabilities (0\%, 20\%, 40\%, 60\%, 80\%) to evaluate the impact of different occlusion levels on vehicle ReID performance.The baseline extracts global features using Vision Transformer(ViT)~\cite{alexey2020image}, TransReID~\cite{he2021transreid} enhances local features via JPM, and VCNet~\cite{zheng2023multi} fuses multi-query features through the VAF module.}
\label{fig:randmask}
\end{figure}

\subsection{Evaluation on random mask query}
% In real-world scenarios, vehicles are often occluded by other vehicles, leading to reduced query accuracy. In such cases, the occlusion caused by foreground vehicles can mislead the model in extracting features of the target vehicle, thereby resulting in incorrect judgments. To address this issue, the proposed CAFNet in this paper effectively reduces the interference from foreground vehicles by complementing and fusing features from multiple perspectives and images. To validate this problem, we designed a data augmentation strategy similar to CutMix~\cite{yun2019cutmix}, where non-target vehicle masks are applied to the original vehicle images with random probabilities, completely occluding the target vehicle. This approach mimics the complex occlusion caused by traffic congestion in real-world environments. The experimental results are shown in Fig.~\ref{fig:randmask}. A single image was randomly selected from the query group as the occlusion object to simulate mutual occlusion between vehicles in real traffic scenarios. In each group, the target vehicle was masked with random probabilities, respectively. This design aims to simulate varying degrees of occlusion and comprehensively evaluate the robustness of the model. 
Real-world vehicle occlusion significantly degrades ReID accuracy by corrupting target feature extraction. To address this, CAFNet leverages multi-view feature complementarity to mitigate occlusion interference. We validate this through a CutMix~\cite{yun2019cutmix}-inspired augmentation strategy that randomly masks target vehicles, simulating varying occlusion levels typical in traffic scenarios. This systematically evaluates model robustness against realistic occlusion conditions.

% The experimental results demonstrate that as the occlusion probability increases, the performance of traditional methods (e.g. Baseline, VCNet and Tansreid) significantly declines. In contrast, the proposed CAFNet effectively reduces occlusion interference and improves ReID accuracy through multi-view feature complementarity and fusion. Specific experimental results are illustrated in Fig.~\ref{fig:randmask}.
Our experiments demonstrate CAFNet's superior robustness against vehicle occlusion compared to existing methods. As shown in Fig.~\ref{fig:randmask}, existing approaches (Vision Transformer(ViT)~\cite{alexey2020image} baseline, TransReID~\cite{he2021transreid}, and VCNet~\cite{zheng2023multi}) suffer significant performance degradation under occlusion, with accuracy dropping sharply as occlusion probability increases. In contrast, CAFNet maintains stable performance through its effective multi-view feature fusion mechanism. The key advantage lies in CAFNet's ability to compensate for occluded regions by leveraging complementary information from multiple perspectives, achieving an average of higher mAP than the best baseline under heavy occlusion conditions. These results validate our method's practical effectiveness in real-world traffic scenarios where inter-vehicle occlusion is common.

% % 视角随机错误
% \begin{table*}[] 
% \caption{ Wrong vehicle images were randomly inserted into the comparison experiment on the LCRI dataset,where $N_q=3$ $reas mask==50\% $}
%     \resizebox{\linewidth}{!}{
%         \begin{tabular}{|l|llll|llll|llll|llll|}
%         \hline
%         random erro  & \multicolumn{4}{c|}{10\%}   & \multicolumn{4}{c|}{25\%}  & \multicolumn{4}{c|}{50\%}  & \multicolumn{4}{c|}{80\%} \\ \hline
%         metric       & mAP  & mCSP & mINP & Rank1  & mAP  & mCSP & mINP & Rank1 & mAP  & mCSP & mINP & Rank1 & mAP  & mCSP & mINP & Rank1 \\ \hline
%         baseline     & 65.6 & 64.0 & 15.2 & 91.8   & 64.0 & 64.1 & 15.0 & 91.0  & 63.0 & 63.2 & 14.3 & 90.4  & 62.6 & 62.0 & 13.1 & 89.5    \\ \hline
%         VCNET        & 50.7 & 49.4 & 11.8 & 72.8   & 47.0 & 45.7 & 10.6 & 68.3  & 39.1 & 38.0 & 8.9  & 56.7  & 39.0 & 37.9 & 9.0  & 55.8  \\ \hline
%         ours         & 69.2 & 67.5 & 17.0 & 93.3   & 68.5 & 66.8 & 16.3 & 92.6  & 66.8 & 65.1 & 15.8 & 91.9  & 65.9 & 64.9 & 14.7 & 90.6  \\ \hline
%         \end{tabular}
%     }
% \label{tab:sota2}
% \end{table*}

% 在实验中，我们设计了5组不同的遮挡概率（0%、20%、40%、60%、80%），以评估不同遮挡程度对车辆检测和再识别（ReID）性能的影响。具体实验设计如下：
% 遮挡对象选择：从查询组中随机选择一张照片作为遮挡对象，模拟真实交通场景中车辆间的相互遮挡现象。
% 遮挡概率设置：在每组实验中，分别以0%、20%、40%、60%、80%的随机概率对目标车辆进行掩码（mask）遮挡。这种设计旨在模拟不同程度的遮挡情况，从而全面评估模型的鲁棒性。
% 方法对比：实验中对比了三种方法：
% Baseline：使用查询图片组的特征均值实现车辆再识别（ReID）。
% Tansreid：同样基于查询图片组的特征均值，但通过多视角特征融合进一步提升性能。
% 实验目标：通过不同遮挡概率下的实验，验证所提出的多查询网络在复杂遮挡场景下的有效性，并分析遮挡程度对检测和再识别精度的影响。
% 实验结果表明，随着遮挡概率的增加，传统方法（如Baseline和Tansreid）的性能显著下降，而本文提出的多查询网络通过多视角特征互补和融合，能够有效降低遮挡干扰，提升检测和再识别的精度。

% VFEM 可视化

\begin{figure}[t]
\centering
    \includegraphics[width=0.8\linewidth]{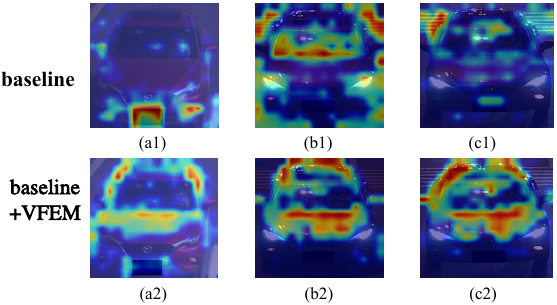}
    \caption{Visualization of the feature maps of our VFEM module comparing with the baseline.}
\label{fig:grad_cvp}
\end{figure}

\begin{figure}[t]
\centering
    \includegraphics[width=1.0\linewidth]{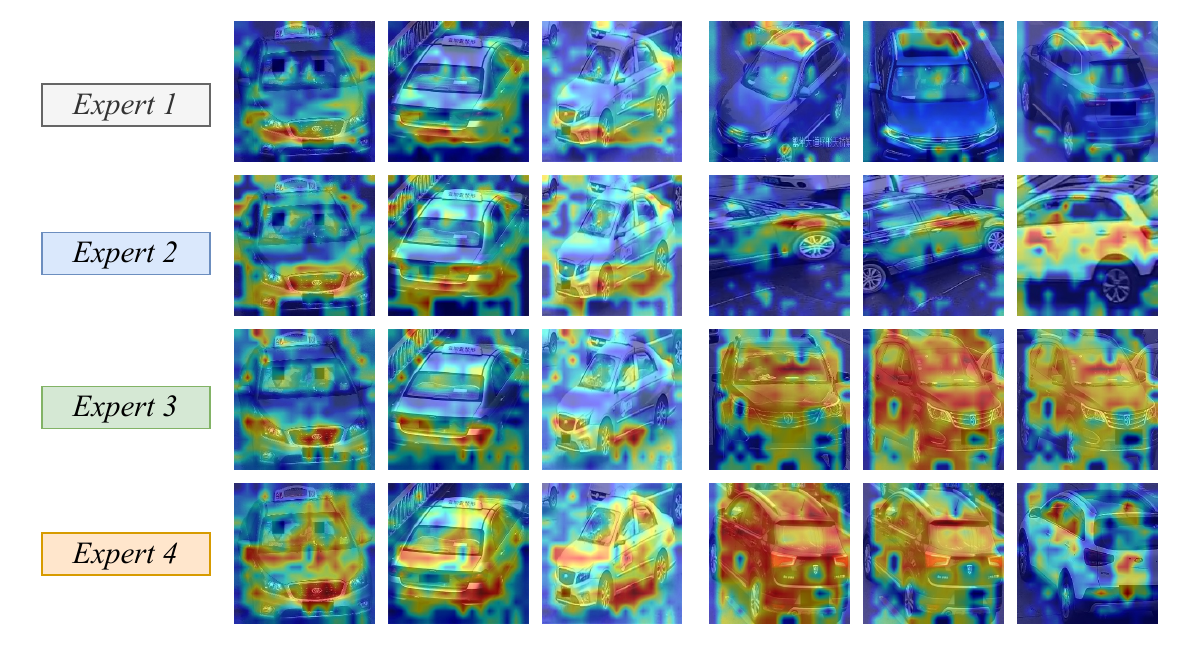}
    \vspace{-4mm}
    \caption{Visualization of the feature maps of our DMFM. Expert attention maps show pronounced region-specific specialization within the vehicle feature space.}
\label{fig:grad_mdf}
\end{figure}

\subsection{Visualization}
To demonstrate the effectiveness of the VFEM module, we visualized the characteristics of the last layer of the model. As shown in Fig.~\ref{fig:grad_cvp}, the VFEM module can effectively reduce background interference and make the model pay more attention to the characteristics of the vehicle itself. For example, (b1) and (b2), (c1) and (c2) in Fig.~\ref{fig:grad_cvp} show the effect of reducing background interference. In addition, the VFEM module also encourages the model to focus on the key features required for classification and explore more areas that are helpful for distinction. This improvement can be seen from the comparison between (a1) and (a2) in Fig.~\ref{fig:grad_cvp}.

Our visualization study randomly selected six vehicle images to examine the DMFM module's behavior. As shown in Fig.~\ref{fig:grad_mdf}, the results reveal that different experts automatically focus on distinct vehicle regions without manual intervention, demonstrating DMFM inherent ability to develop specialized attention patterns for effective multi-view fusion. This random sampling approach confirms the generalizability of the module's adaptive feature integration capability across diverse vehicle appearances.

\section{Conclusion}
\label{sec:conclusion}
% 本文提出了一种新颖的跨摄像头多查询车辆重识别网络 (CAFNet)，旨在高效整合来自不同视角的车辆特征信息。通过设计跨视角感知模块（VFEM）和视角动态融合模块（MDF），该网络利用成对交叉注意力机制提取互补信息，从而生成更具辨别力和丰富的特征。同时，MDF模块自适应调整特征权重，实现原始特征与增强特征的动态融合，最终生成统一的全局特征表示，以提高车辆重识别性能。
% In this paper, we proposes Cross-view Adaptive Fusion Network (CAFNet) that aims to efficiently integrate vehicle feature information from different viewpoints. First, we propose a Cross-view Invariant Perception (VFEM) module in the training and inference process to learn viewpoint invariant information.
% Then, we propose a Multi-view Dynamic Fusion (MDF) module to integrate the complementary information among different viewpoints in the training and inference process.
% More over, we proposes a Multi-view Alignment Loss. MAL achieves feature alignment through bidirectional cross-view contrastive learning and reconstruction constraints.
% Finally, a new dataset (LCRI-1K) are proposed to measure the ability of cross-scene recognition and conduct multi-query evaluation experiments respectively.
% Comprehensive experiments demonstrate the necessity of the effectiveness of the proposed CAFNet.
% This work provides new research direction for vehicle Re-ID and related areas.

This paper proposes a Cross-view Adaptive Fusion Network (CAFNet) for multi-query vehicle re-identification, which effectively addresses the limitations of existing methods in modeling view relationships and feature fusion through three key technical innovations. First, the proposed View-specific Features Enhancement sub-Module (VFEM) module captures complementary information across different views through a pairwise cross-attention mechanism. Second, the designed Dynamic Multi-view Fusion sub-Module (DMFM) achieves view-aware adaptive feature fusion based on a mixture-of-experts mechanism. Additionally, the introduced Multi-view Alignment Loss (MAL) ensures feature consistency through bidirectional cross-view contrastive learning and reconstruction constraints. To comprehensively evaluate algorithm performance in real-world complex scenarios, we constructed LCRI-1K. Extensive experiments demonstrate that CAFNet shows significant advantages in challenging scenarios involving occlusion and viewpoint variations. This research provides a novel solution for multi-query vehicle re-identification task.

%------------------------------------------------------------------------
%%%%%%%%% REFERENCES
\bibliographystyle{IEEEtran}
\small\bibliography{egbib}

\end{document}